\documentclass{article}

\PassOptionsToPackage{square,numbers}{natbib}

\usepackage[preprint]{neurips_2021}

\usepackage[utf8]{inputenc} 
\usepackage[T1]{fontenc}    
\usepackage{hyperref}       
\usepackage{url}            
\usepackage{booktabs}       
\usepackage{amsfonts}       
\usepackage{nicefrac}       
\usepackage{microtype}      
\usepackage{xcolor}         

\usepackage{listings}
\usepackage{algorithmic}
\usepackage{algorithm}
\usepackage{booktabs}
\usepackage{amsmath}
\usepackage{amssymb}
\usepackage{bm}
\usepackage{subcaption}
\usepackage{multirow}
\usepackage{tikz}
\usetikzlibrary{shapes, arrows}
\tikzstyle{circlenodeshaded} = [draw, fill={rgb:black, 1; white, 8}, circle]
\tikzstyle{circlenodelatent} = [draw, circle]

\title{Rethinking Noisy Label Models: Labeler-Dependent Noise with Adversarial Awareness}

\author{%
  Glenn Dawson \\
  Rowan University \\
  Glassboro, New Jersey, USA \\
  \texttt{dawsong@rowan.edu}
  \And
  Robi Polikar \\
  Rowan University \\
  Glassboro, New Jersey, USA \\
  \texttt{polikar@rowan.edu}
}

\begin{document}

\maketitle

\begin{abstract}
Most studies on learning from noisy labels rely on unrealistic models of i.i.d.\@ label noise, such as class-conditional transition matrices.
More recent work on instance-dependent noise models are more realistic, but assume a single generative process for label noise across the entire dataset.
We propose a more principled model of label noise that generalizes instance-dependent noise to multiple labelers, based on the observation that modern datasets are typically annotated using distributed crowdsourcing methods.
Under our labeler-dependent model, label noise manifests itself under two modalities: \textit{natural error} of good-faith labelers, and \textit{adversarial labels} provided by malicious actors.
We present two adversarial attack vectors that more accurately reflect the label noise that may be encountered in real-world settings, and demonstrate that under our multimodal noisy labels model, state-of-the-art approaches for learning from noisy labels are defeated by adversarial label attacks.
Finally, we propose a multi-stage, labeler-aware, model-agnostic framework that reliably filters noisy labels by leveraging knowledge about which data partitions were labeled by which labeler, and show that our proposed framework remains robust even in the presence of extreme adversarial label noise.
\end{abstract}

\section{Introduction}
Learning from labeled data relies on the availability of accurate labels.
However, obtaining accurate labels on large datasets is costly, sometimes prohibitively so \cite{KARIMI2020101759, esteva2017dermatologist, gulshan2016development, li2019learning}, which has led to crowdsourcing as an attractive and cost-effective solution for distributed label gathering \cite{zhang2017improving, raykar2009supervised, whitehill2009whose, welinder2010multidimensional, zhong2017quality, sheng2017majority, schroff2010harvesting}. 
Unfortunately, the labels obtained via crowdsourcing are of unreliable veracity, and models trained naively on such unreliable data are vulnerable to overfitting on noisy labels \cite{45820, NEURIPS2018_f2925f97}.
Furthermore, some labelers on crowdsourcing platforms may be adversarial, providing intentionally incorrect labels and greatly increasing the incidence of label noise \cite{jagabathula2017identifying, pmlr-v80-kleindessner18a, ma2020adversarial, checco2020adversarial}.

Learning in the presence of noisy labels has emerged as an area of active research \cite{frenay2013classification, algan2021image, song2020learning}, and generally follows one of two modalities.
The \textit{learning from crowds} approach assumes that labels are generated by multiple annotators, each with an associated but unknown probability of generating a correct label \cite{donmez2010probabilistic, raykar2010learning, yan2014learning, zhang2014spectral, donmez2009efficiently}, while the \textit{learning from noisy labels} approach treats the label noise as i.i.d.\@, with an unobserved label flipping process governing the incidence of label noise for all examples \cite{NIPS2013_3871bd64, bekker2016training, yi2019probabilistic, arazo2019unsupervised}.
While both approaches have been successful for addressing label uncertainty under their respective assumptions, they each depend on unrealistic restrictions.
Methods for learning from crowds rely on redundant labels, with the ``wisdom of the crowd'' overruling unreliable or adversarial labels \cite{roman2009crowdsourcing, zhang2016learning, vaughan2017making}. 
However, requiring multiple labels per data instance greatly increases the burden of label gathering, running counter to the original problem of label gathering.
Meanwhile, both class-conditional and instance-dependent approaches for learning from noisy labels ignore important information about the source of each label that can be leveraged for more robust performance, as well as the possibility of non-i.i.d.\@ label noise distributions \cite{hickey1996noise, frenay2013classification}.

We address these drawbacks by combining the strengths of each approach.
We propose a labeler-dependent model of label noise that inherits the multiple-labeler paradigm of crowdsourcing \textit{without} requiring overlapping label redundancy, while simultaneously incorporating instance-dependent noise modeling and introducing the threat of adversarial labels.
We believe that our framework more accurately models the types of label noise that a learner may encounter in a real-world scenario.

Furthermore, to address the challenges introduced by our multimodal label noise model, we propose a novel three-stage learning framework.
We draw inspiration from social learning theory as well as theory of mind, which provide evidence for humans modeling the unobserved opinions of others during their own learning and decision-making processes \cite{kovacs2010social, frith2012role, park2017integration, joiner2017social, khalvati2019modeling, da2020humans}.
First, we create an ensemble of semi-supervised models of each labeler, and query each labeler model for an estimate of the label that its corresponding labeler \textit{would} have provided for each data instance, thus achieving synthetic label redundancy without requiring overlapping labels directly from each labeler.
Then, we integrate the queried labels into a single label per instance that is more reliable than the queried labels using methods for learning from crowds.
Finally, we use our filtered (but still noisy) labels to train a classifier using algorithms for learning from noisy labels.
Our framework is modular and agnostic to its component algorithms, allowing for refinement in each stage as respective advances continue in the fields of semi-supervised learning, learning from crowds, and learning from noisy labels.


\section{Preliminaries}
\label{sec:preliminaries}
For a $K$-class problem, we define the label space $\mathbf{Y}$ as a $K \times K$ identity matrix for one-hot encoding of the classes, with $y_k$ representing the $k^{\mathrm{th}}$ class label.
The true label for instance $\bm{x}_i$ is denoted as $y_i \in \mathbf{Y}$, and a noisy label for instance $\bm{x}_i$ is denoted as $\hat{y}_i \in \{\mathbf{Y} \setminus y_i\}$.

\paragraph{Class-conditional label noise}
Label noise is commonly treated as a class-conditional phenomenon, where the noisy labels are treated strictly as a function of the true label \cite{ma2018dimensionality, 48960, reeve2019classification, NEURIPS2020_f4e3ce3e, NEURIPS2020_e0ab531e, NEURIPS2020_512c5cad, liu2020early, NEURIPS2020_c6102b37, NEURIPS2020_8493eeac, angluin1988learning, chen2019understanding, NEURIPS2019_8cd7775f, pmlr-v89-amid19a, wei2020optimizing, xiao2015learning, thulasidasan2019combating, yu2019does, wang2020training}.
Under this treatment, label corruption is modeled using a transition matrix $T$ describing the probability of a true label being randomly flipped \cite{NIPS2013_3871bd64, patrini2017making, cheng2020learning, tanno2019learning}.
Two common forms of $T$ for performing evaluations of noisy label learning include \textit{symmetric} and \textit{asymmetric} noise \cite{tanaka2018joint}.
Under symmetric noise, $T$ follows a uniform distribution, with each label $y_i$ having an equal probability of corruption to any other label $y_k$. 
In contrast, under asymmetric label noise $T$ is deliberately constructed to place higher probabilities on classes heuristically similar to $y_i$ \cite{patrini2017making, scott2013classification, li2019learning, sukhbaatar2015training, ren2018learning, NEURIPS2018_a19744e2, wang2021tackling}.


\paragraph{Instance-dependent label noise}
Both symmetric and asymmetric label noise are attractive schemes for synthetically generating label noise, as they allow complete control over the amount of label noise injected into the dataset.
However, these approaches ignore the data-driven dependencies of the labels, and produce unrealistic data-label distributions. 
More realistic models include early work on feature-dependent label noise \cite{lachenbruch1974discriminant, manwani2013noise}, and, more recently, instance-dependent noise \cite{garcia2019new, chen2020beyond, wang2021tackling, cheng2020learning, zhu2020second, zhang2021learning, NEURIPS2020_5607fe88}. 
One such approach involves projecting each instance onto a randomly-sampled set of $K$ vectors and sampling from a combination of this projection with the clean label to generate a noisy label \cite{NEURIPS2020_5607fe88, zhu2020second, cheng2020learning}.
Other approaches generate instance-specific noise by training a deep neural network on the clean training dataset and using its noisy outputs; some authors use the entire set of noisy labels generated by the network \cite{wang2021tackling}, while others retain only a percentage of the noisy labels \cite{chen2020beyond}.
The polynomial margin diminishing noise introduced in \cite{zhang2021learning} extends confidence-based noise to stochastically flip labels based on the outputs of a neural network trained on clean labels.

\paragraph{Noisy labels from crowdsourcing}
Both class-conditional and instance-dependent noise models assume that the noisy labels are generated by a single process for the entire training dataset. 
This is an unrealistic assumption in the context of big data, where labels may be gathered via distributed methods such as crowdsourcing. 
Addressing unreliable labels is a fundamental objective of learning from crowds \cite{snow2008cheap, ensemble-linear, zhou2014aggregating, rodrigues2013learning, zhang2018multi, chen2013optimistic, NIPS2012_cd00692c, ghosh2011moderates, dalvi2013aggregating, tao2020label}, and can be traced to the seminal work of Dawid and Skene, who modeled the accuracy of each labeler as a hidden confusion matrix \cite{dawid1979maximum}.
The GLAD model adds instance-specific dependence by modeling the difficulty of correctly labeling each instance alongside each labeler's expertise \cite{whitehill2009whose}.
A parameterization assuming that labels are generated from a Gaussian mixture model encoding both the labeler- and instance-specific noise was proposed in \cite{welinder2010multidimensional}.
The personal classifiers model assumes that labels are generated by labeler-specific logistic regression models, characterized by their deviation from an optimal base classifier \cite{kajino2012convex}.


\paragraph{Adversarial label noise}
Despite a considerable body of work on adversarial evasion attacks against deep neural networks, there is comparatively little research on adversarial label attacks.
The possibility of adversarial labelers in crowdsourcing was acknowledged in \cite{whitehill2009whose}, but without explicitly addressing such threats. 
Other efforts have investigated the vulnerability of support vector machines to label flipping attacks \cite{biggio2011support}, and have studied identifying adversarial workers in crowdsourcing \cite{jagabathula2017identifying}.  
Arbitrary adversaries that violate the Dawid-Skene model, which may even be in collusion, were considered in \cite{pmlr-v80-kleindessner18a}; their work was extended to the task assignment matrix not being dense in \cite{ma2020adversarial}. 

\section{Labeler-dependent noise: mixing natural error and adversarial labels}
\label{sec:multimodal-label-noise}
\begin{figure}
    \centering
    \begin{subfigure}{.5\textwidth}
        \centering
        \label{models-a}
        \includegraphics[width=.9\linewidth]{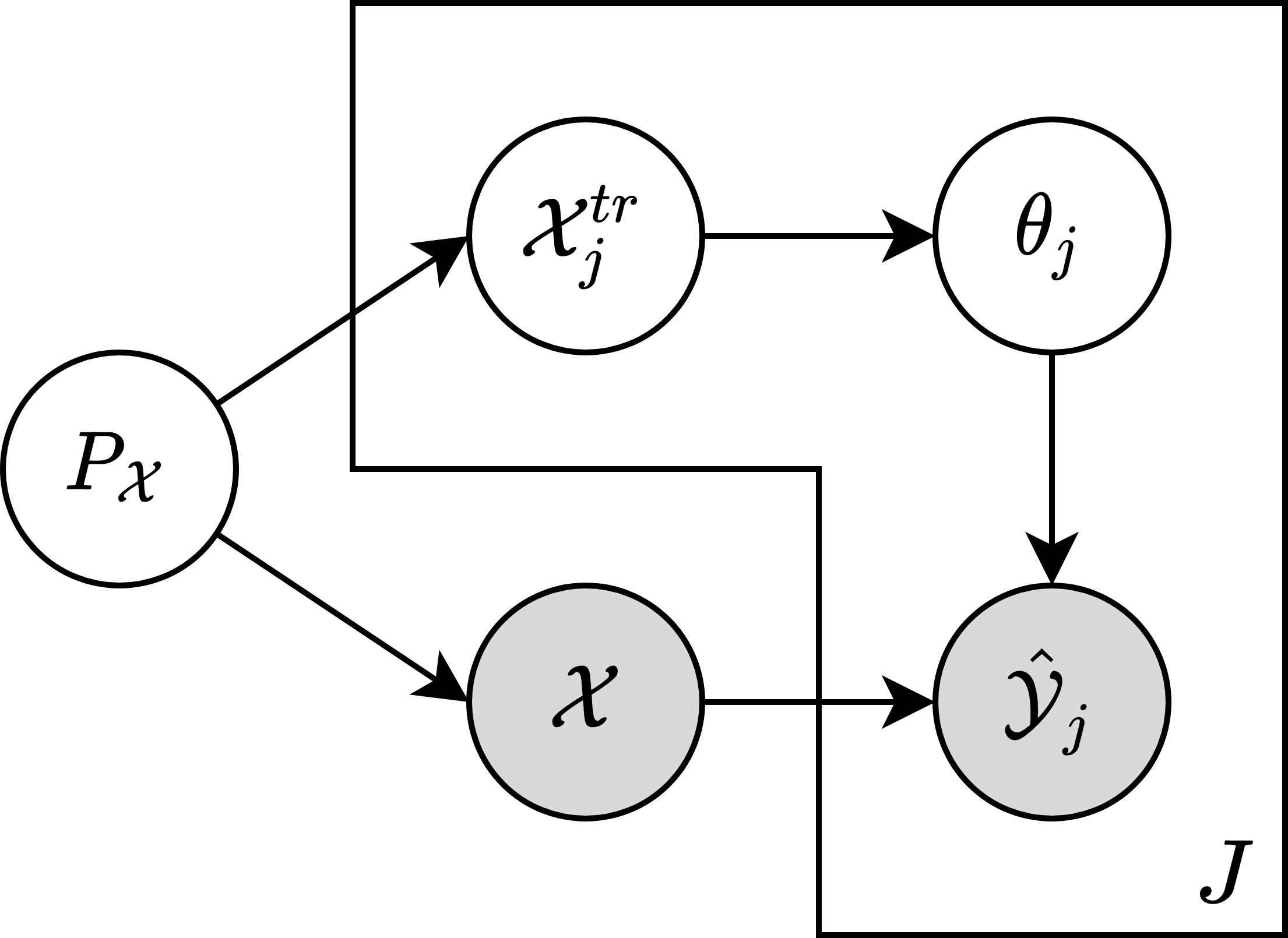}
    \end{subfigure}\hfill
    \begin{subfigure}{.5\textwidth}
        \centering
        \includegraphics[width=.9\linewidth]{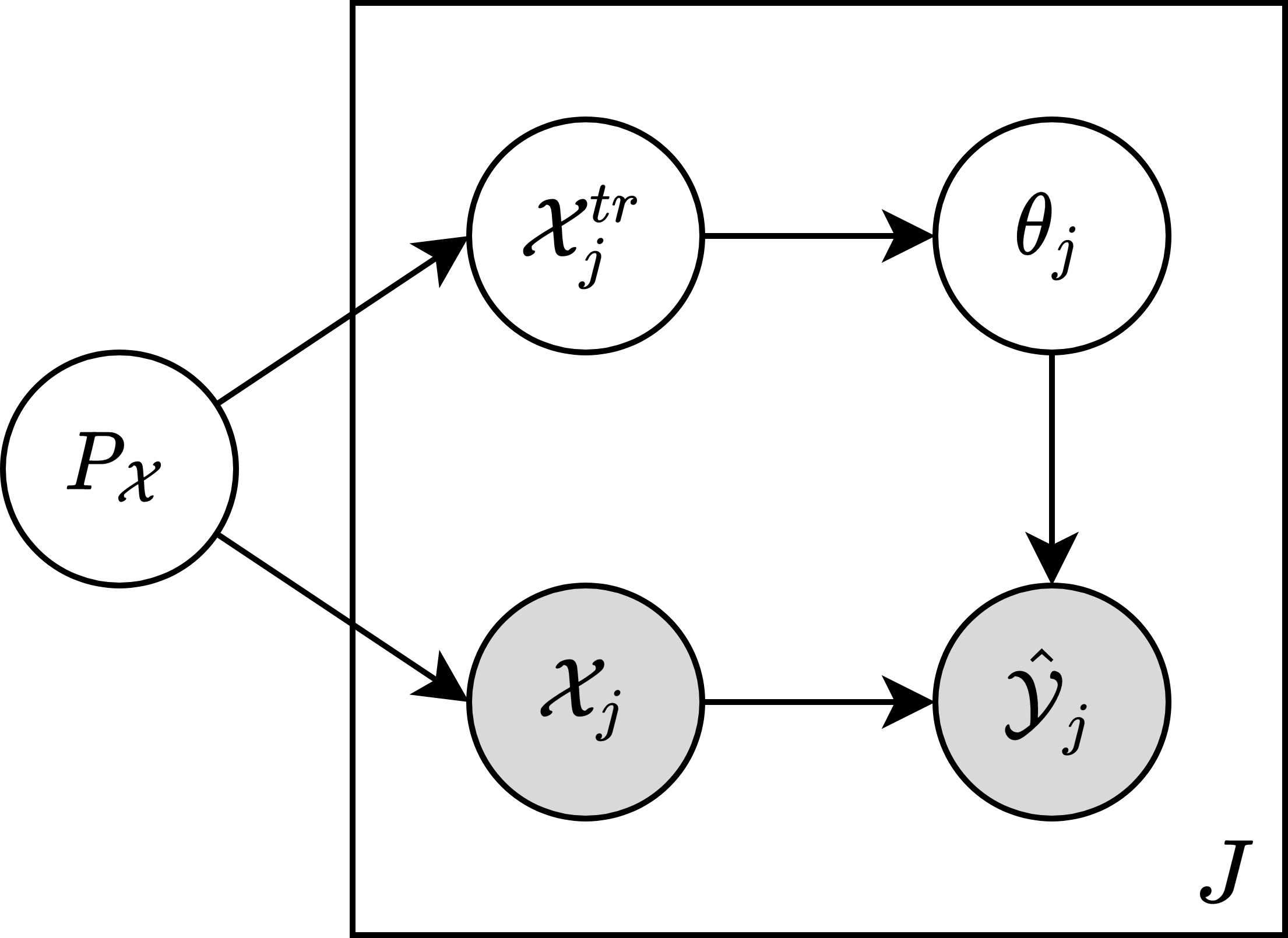}
    \end{subfigure}
    \caption{
        \textbf{Left:} In conventional crowdsourcing, the full training dataset $\mathcal{X}$ is drawn from the underlying distribution $P_\mathcal{X}$, and $J$ labelers provide redundant labels $\hat{\mathcal{Y}}_j$ on $\mathcal{X}$. 
        \textbf{Right:} Our model assumes that each labeler $j$ observes an independently-drawn subset $\mathcal{X}_j \sim P_\mathcal{X}$, and provides labels $\hat{\mathcal{Y}}_j$ only on $\mathcal{X}_j$. 
        In both models, the parameters $\theta_j$ are trained on independently-drawn data $\mathcal{X}^{tr}_j \sim P_{\mathcal{X}}$.
    }
    \label{fig:crowdsourcing-model}
\end{figure}

In this section, we propose a hybrid labeler-dependent model for noisy labels that integrates key insights from each of the previously-discussed settings.
To our knowledge, ours is the first work that has attempted to extend instance-dependent label noise to the multiple-labelers paradigm.

\paragraph{Multiple labelers}
We introduce the multiple-labelers paradigm from crowdsourcing into the instance-dependent noise model by modeling the training dataset $\mathcal{X}$ as the concatenation of multiple partitioned subsets $\mathcal{X}_j$, each labeled by an independent labeler $j$.
We note the differences between our proposed model and the conventional crowdsourcing model, which assumes that all labelers provide redundant labels on the same dataset $\mathcal{X}$; instead, we consider that each data instance is associated with only a single label.
Assuming that each $\mathcal{X}_j$ is drawn from the same underlying distribution $P_{\mathcal{X}}$, the resultant dataset $\mathcal{X} = \{\bigcup_j \mathcal{X}_j\}$ is identical to a supervised dataset with noisy labels, but with additional information about which subsets of data were labeled by which labeler. 
The differences between the conventional crowdsourcing model and our proposed model are shown in Figure \ref{fig:crowdsourcing-model}.

\paragraph{Natural error of good-faith labelers}
We consider each labeler $j$ to generate labels following a categorical distribution parameterized by $\theta_j$ such that $\hat{y}_{ij} \sim \mathrm{Cat}(\mathbf{Y}|\bm{\pi}_{ij})$, where $\bm{\pi}_{ij} = f_{\bm{\theta}_j}(\bm{x}_i)$ can be thought of as the softmax output of the classifier $f_{\theta_j}$ with $\theta_j$ learned on an unobserved training dataset $\mathcal{D}_j^{tr} = \{\mathcal{X}_j^{tr} \nsubseteq \mathcal{X}, \mathcal{Y}_j^{tr}\}$. 
The natural classification error of each labeler assumes the role of instance-dependent label noise; our model is similar to the instance-dependent noise technique of training a deep neural network on the cleanly-labeled training data and using its outputs as noisy labels.
However, we distinguish our model in two ways: (i) the noisy labels are generated not by a single neural network but by an ensemble of independent networks, causing more complex instance- and labeler-dependent label noise dynamics; and (ii), the parameters $\theta_j$ are not learned on the same data for which the noisy labels are provided, avoiding data leakage and maintaining a more realistic model of the relations between the labelers and the data.

\paragraph{Adversarial labels from bad-faith labelers}
The good-faith labeler model can be extended to describe \textit{adversarial} labelers. 
We posit that an intelligent adversary will present the learner with a false label based on not only its own best guess as to the correct label, but also its \textit{second} best guess.
We define the function $\mathrm{arg}_2\mathrm{max}$ as identical to the $\mathrm{argmax}$ function, except that $\mathrm{arg}_2\mathrm{max}(A)$ returns the index of the element of $A$ with the second highest value instead of the maximum (with ties broken arbitrarily).
An adversarial labeler therefore provides noisy labels following
\begin{equation}
\label{eqn:label-theta-pi-adv}
    \hat{y}_{ij} = \mathrm{arg}_2\mathrm{max}_k\ f_{\theta_j}(\bm{x}_i).
\end{equation}  

The labels that the adversary presents are representative of the adversary's best guess about the false labels that are most likely to be confused with the correct labels.
This approach is similar to that in \cite{zhang2021learning}, where the label of the second most-confident category is integrated into the noise model.
Note that the adversary may itself have an incorrect belief about the true label, and in its attempt to provide a false label may accidentally provide a correct label.

Following the crowdsourcing paradigm, we assume that the label gathering process draws from a pool of labelers containing a mixture of both good-faith (but imperfect) labelers and malicious, adversarial labelers.
By combining natural and adversarial noise from multiple labelers, we define a multimodal, labeler-dependent model that is both more realistic than class-conditional approaches and more general than single-process instance-dependent approaches.
We note that the characteristics of both natural and adversarial noise will vary from labeler to labeler, and that even when adversarial noise is absent there may be variations in the levels and characteristics of the natural error associated with each labeler.

\paragraph{Multiple-labeler adversarial attack vectors}
We define two realistic vectors of adversarial attacks that should form the basis of future work on adversary-aware learning from noisy labels.
A \textit{data flooding attack} occurs when a single adversary provides an overwhelming quantity of adversarial labels relative to the quantity of labels provided by the good-faith labelers.
A \textit{multiple adversaries attack} occurs when multiple bad-faith labelers invade the labeling process; unlike the data flooding attack, each adversarial labeler need not provide large quantities of labels.
Both attacks exploit vulnerabilities in distributed label collection, especially in crowdsourcing and online learning scenarios, and have the capability to introduce large amounts of adversarial noise into the training data.  

Under our model, an algorithm for learning from noisy labels must be resilient to both natural and adversarial label noise, and algorithmic robustness can be measured as its vulnerability to increasing levels of adversarial labels.
While the most favorable learning environment will include no adversarial noise, a robust algorithm will exhibit little to no reduction in performance as more noise is introduced; a weak algorithm, however, will be heavily compromised by increasingly adversarial labels, and its performance will be negatively impacted. 

\section{A three-stage framework for labeler- and adversary-aware learning}
\label{sec:sslfc-outline}

\begin{algorithm}
    \caption{Stage one: semi-supervised labeler modeling.}
    \begin{algorithmic}[1]
        \label{alg:ssl-lfc}
        \renewcommand{\algorithmicrequire}{\textbf{Inputs:}}
        \renewcommand{\algorithmicensure}{\textbf{Output:}}
        \REQUIRE $\mathcal{D} = \{\mathcal{D}_1, \dots, \mathcal{D}_J\}$, the set of data-label pairs provided by each label source $j \in J$; $\mathrm{SSL}$, a semi-supervised classification algorithm
        \STATE $\bm{\Phi} \leftarrow $ Initialize an empty set of trained models
        \FOR {each labeler $j=1$ to $J$}
            \STATE $\mathcal{X}_L, \hat{\mathcal{Y}}_L \leftarrow \mathcal{D}_j$  (Gather labeled data from labeler $j$)
            \STATE $\mathcal{X}_U \leftarrow \{\bigcup_{\ell \neq j} \mathcal{X}_\ell\}\ \forall \ell \in J$ (Gather unlabeled data from all other labelers $\ell \neq j$)
            \STATE $\bm{\phi}_j \leftarrow \mathrm{SSL}(\mathcal{X}_L, \hat{\mathcal{Y}}_L, \mathcal{X}_U)$ (Estimate $\theta_j$)
            \STATE $\bm{\Phi} \leftarrow \{\bm{\Phi} \cup \bm{\phi}_j\}$
        \ENDFOR
        \ENSURE $\bm{\Phi}$, the set of estimates of labeler parameters
    \end{algorithmic}
\end{algorithm}
We consider the set of partitioned subsets of noisily-labeled data provided by each labeler, $\mathcal{D} = \{\mathcal{D}_1, \dots, \mathcal{D}_J\}$, where each $\mathcal{D}_j = \{\mathcal{X}_j, \hat{\mathcal{Y}}_j\}$.
Because each labeler $j$ provides labels only on $\mathcal{X}_j$, and each $\mathcal{X}_j$ is disjoint from every other $\mathcal{X}_{\ell \neq j}$, we cannot directly use methods for learning from crowds to exploit our labeler-aware knowledge.
To accommodate this scenario, Stage 1 of our framework trains a set of models $\bm{\Phi} = \{\phi_1, \dots, \phi_J\}$, respectively estimating each $\theta_j$, which can then be queried to obtain \textit{estimates} of the labels that each labeler $j$ \textit{might} have provided.
To do so, we observe that while the labels $\hat{\mathcal{Y}}_{\ell \neq j}$ from all labelers $\ell \neq j$ are uninformative for estimating $\theta_j$, the data $\mathcal{X}_{\ell \neq j}$ may still be exploited assuming that each $\mathcal{X}_j$ are drawn from the same underlying distribution $P_{\mathcal{X}}$.
We leverage semi-supervised learning (SSL) to train each $\phi_j$ by treating the data-label pairs $\{\mathcal{X}_j, \hat{\mathcal{Y}}_j\}$ as labeled data, and the union $\{\bigcup_\ell \mathcal{X}_{\ell \neq j}\}$ as unlabeled data.
For modularity, we place no restrictions on which SSL methods may be used.
It is also not essential for any $\phi_j$ to achieve high accuracy on test data; in this stage, we are interested only in estimating, as closely as possible, the parameters $\theta_j$ that describe the label generation dynamics of each labeler $j$, regardless of generalization performance.
The procedure for semi-supervised labeler modeling is shown in Algorithm \ref{alg:ssl-lfc}.

\begin{algorithm}
    \caption{Stage two: combining real and synthetic labels with learning from crowds.}
    \begin{algorithmic}[1]
        \label{alg:ssl-lfc-crowd}
        \renewcommand{\algorithmicrequire}{\textbf{Inputs:}}
        \renewcommand{\algorithmicensure}{\textbf{Output:}}
        \REQUIRE $\mathcal{D} = \{\mathcal{D}_1, \dots, \mathcal{D}_J\}$, the set of data-label pairs provided by each label source $j \in J$; $\bm{\Phi} = \{\bm{\phi}_1, \dots, \bm{\phi}_J\}$, the set of estimates of each $\theta_j$ for each label source $j$; $\mathrm{LFC}$, an algorithm for classification by learning from crowds
        \STATE $\mathcal{W} \leftarrow $ Initialize an empty $J \times N$ matrix of labels from $J$ labelers on $N$ instances
        \FOR {each labeler $j = 1$ to $J$}
            \FOR {each dataset $\ell = 1$ to $J$}
            \STATE $\mathcal{W}_{j\ell} \leftarrow \begin{cases}
                \hat{\mathcal{Y}_j}, & \ell = j\\
                g_{\phi_j}(\mathcal{X}_\ell), & \ell \neq j
            \end{cases}$
            \ENDFOR
        \ENDFOR
        \STATE $\widetilde{\mathcal{Y}} \leftarrow \mathrm{LFC}(\mathcal{W})$ (Estimate true labels from labels matrix using learning from crowds)
        \ENSURE $\widetilde{\mathcal{Y}} $, the set of estimates for the true labels of $\mathcal{X}$
    \end{algorithmic}
\end{algorithm}

Once all $\phi_j$ are obtained,  Stage 2 of our framework queries each model to obtain synthetic labels on the entire set of training data.
Combining these synthetic labels (queried from $\phi_j$) with the labeler-provided labels $\hat{\mathcal{Y}}_j$ (generated by $\theta_j$), we now have a full set of $J$ redundant labels for each $\bm{x}_i \in \mathcal{X}$.
Hence, we can use learning from crowds to integrate the redundant noisy labels into a single filtered label per instance.
As in Stage 1, we place no restrictions on which methods for learning from crowds may be utilized.
Combining real and synthetic labels is shown in Algorithm \ref{alg:ssl-lfc-crowd}.

\begin{algorithm}
    \caption{Three-stage framework for labeler-aware learning under adversarial label noise.}
    \begin{algorithmic}[1]
        \label{alg:three-stage}
        \renewcommand{\algorithmicrequire}{\textbf{Inputs:}}
        \renewcommand{\algorithmicensure}{\textbf{Output:}}
        \REQUIRE $\mathcal{D} = \{\mathcal{D}_1, \dots, \mathcal{D}_J\}$, the set of data-label pairs provided by each label source $j \in J$; $\mathrm{SSL}$, a semi-supervised classification algorithm; $\mathrm{LFC}$, an algorithm for classification by learning from crowds; $\mathrm{LNL}$, an algorithm for learning from noisy labels
        \STATE Learn $\bm{\Phi}(\mathcal{D}, \mathrm{SSL})$ following Algorithm \ref{alg:ssl-lfc}
        \STATE Learn $\widetilde{\mathcal{Y}}(\mathcal{D}, \bm{\Phi}, \mathrm{LFC})$ following Algorithm \ref{alg:ssl-lfc-crowd} to filter adversarial noise
        \STATE Learn $\Theta \leftarrow \mathrm{LNL}(\widetilde{\mathcal{Y}})$ to filter natural noise
        \ENSURE $\Theta$, a classifier for learning from multiple noisy labelers
    \end{algorithmic}
\end{algorithm}

While methods for learning from crowds are effective at filtering adversarial noise, the labels that they produce are not free of natural noise.
To reduce the impact of natural noise, we use the labels produced in Stage 2 to train $\Theta$, a model for learning from noisy labels.
The first two stages therefore constitute a bootstrapping procedure (similar to \cite{reed2014training}), with Stage 3 being stacked on top of the bootstrapped labels.
As before, we place no restrictions on which methods for learning from noisy labels can be used.
The complete procedure for our three-stage framework is shown in Algorithm \ref{alg:three-stage}.

\section{Comparisons to state-of-the-art under adversarial label noise}
\label{sec:experiments}
We demonstrate our label-gathering process, as well as our proposed algorithm for addressing multimodal label noise, via experiments that simulate the process under both data flooding and multiple adversaries attack scenarios. 
We used the MNIST \cite{lecun-mnisthandwrittendigit-2010} and SVHN \cite{svhn} datasets, two commonly-used benchmark datasets for learning from noisy labels.
We evaluate against the DivideMix \cite{li2019dividemix}, self-evolution average label (SEAL) \cite{chen2020beyond}, and progressive label correction (PLC) \cite{zhang2021learning} algorithms, representing the current state-of-the-art in learning from noisy labels under both class-conditional and instance-dependent noise models.
We also compare the effectiveness of our framework using only the first two stages against the complete three-stage framework.

\paragraph{Experimental setup}
Our labeler-dependent noise model generates noisy labels as a function of $\theta_j$, preventing precise control of the amount of label noise. 
Instead, we parameterize each experiment by the amount of data provided by adversarial labelers.
For data flooding attack, we fix the amount of data provided by good-faith labelers and vary the amount of data provided by a single adversarial labeler; for the multiple adversaries attack, we fix the number of labelers and the amount of data provided by each labeler, and vary the number of adversarial labelers.

For each experiment, we modeled $J$ label sources  by training $J$ neural networks $\theta_j$ on small subsets $\mathcal{D}^{tr}_j$, drawn without replacement from the training dataset $\mathcal{D}$. Each labeler $j$ then provided labels on randomly-partitioned subsets $\mathcal{X}_j$ of the remainder of the training data, with $\bigcup_j \mathcal{X}_j = \{\mathcal{X} \setminus \{\bigcup_j \mathcal{X}^{tr}_j\}\}$. 
Good-faith labelers provided labels following $\hat{y}_{ij} = \mathrm{argmax}_k\ f_{\theta_j}(\bm{x}_i)$, while adversarial labelers followed $\hat{y}_{ij} = \mathrm{arg}_2\mathrm{max}_k\ f_{\theta_j}(\bm{x}_i)$ (Eqn.\@ \ref{eqn:label-theta-pi-adv}).
For the MNIST dataset, we set $J = 10$, $N_{tr} = 200$, and each $\theta_j$ was a randomly-initialized ResNet-18 \cite{he2016deep}, producing  \textasciitilde 7.6\% natural error. 
For the SVHN dataset, we set $J = 5$, $N_{tr} = $ 20,000, and each $\theta_j$ was a randomly-initialized Wide ResNet-50 \cite{Zagoruyko2016WRN}, producing \textasciitilde 9\% natural error.
The complete experimental procedure is detailed in Appendix C.

Once the noisy labels were generated by each labeler, the noisily-labeled data were passed to the learning algorithm under test. 
For our labeler-aware approach, we provided each labeler's data-label pairs  $\{\mathcal{X}_j, \hat{\mathcal{Y}}_j\}$ as separate datasets. 
For the labeler-agnostic approaches, we combined the data-label pairs from each source into a single dataset (i.e.\@ as they would be observed under labeler-agnostic assumptions).
We repeated each experiment ten times for MNIST and five times for SVHN, and we report the results as the mean classification accuracies in response to increasing adversarial noise, bounded by their 95\% confidence intervals based on the two-sided Student's $t$-test.

\paragraph{Model selection and hyperparameter tuning}
Due to modular nature of our framework, the selections for the semi-supervised learning, learning from crowds, and learning from noisy labels algorithms to use for each stage constitute its main high-level hyperparameters.
We note that our framework's modularity allows for the seamless replacement of any or all of these algorithmic choices; we demonstrate this modularity by selecting different SSL algorithms for each experiment.
For the MNIST dataset, we use auxiliary deep generative models as our SSL algorithm due to its small parameter footprint \cite{maaloe2016auxiliary}; for the SVHN dataset, we chose the FixMatch algorithm as representative of the current state-of-the-art for semi-supervised learning \cite{NEURIPS2020_06964dce}.
For both datasets, we use OpinionRank as our learning from crowds algorithm due to its nonparametric nature and fast performance \cite{dawson2021opinionrank}, and selected DivideMix for learning from noisy labels due to its state-of-the-art performance on a wide variety of noisy labels tasks \cite{li2019dividemix}.

Since we cannot assume the presence of a clean validation set, we do not perform any ground truth-based hyperparameter tuning or model selection, as in \cite{arazo2019unsupervised}.
Instead, the hyperparameters for each component of our modular framework, as well as those of the algorithms against which we are comparing, were selected based on the suggestions of the original authors of each algorithm, available in their publicly-available online implementations.
We made only minor adjustments in order to accommodate differences in intended datasets, and in all cases we tested our changes in the non-adversarial setting to ensure fair comparison.
All implementation details and experimental parameters can be found in Appendix C.

Due to lack of any fine-tuning, we are likely reporting conservative results for all algorithms, including our own.
However, we are emphatically \textit{not} presenting our results as benchmark scores; rather, we are interested in the overall \textit{trends of behavior} that characterize the vulnerabilities of each algorithm to increasing intensities of adversarial attacks.
While it may be possible to obtain minor improvements through extensive hyperparameter tuning on a clean validation dataset, we argue that because we cannot assume the existence of such a set, such tuning would constitute unfair data leakage.
More importantly, we believe that such minor improvements would not change the overall structure of our results with respect to the characteristics of each algorithm's vulnerability to adversarial label noise.

\paragraph{Data flooding}
\begin{figure}
    \centering
    \includegraphics[width=\linewidth]{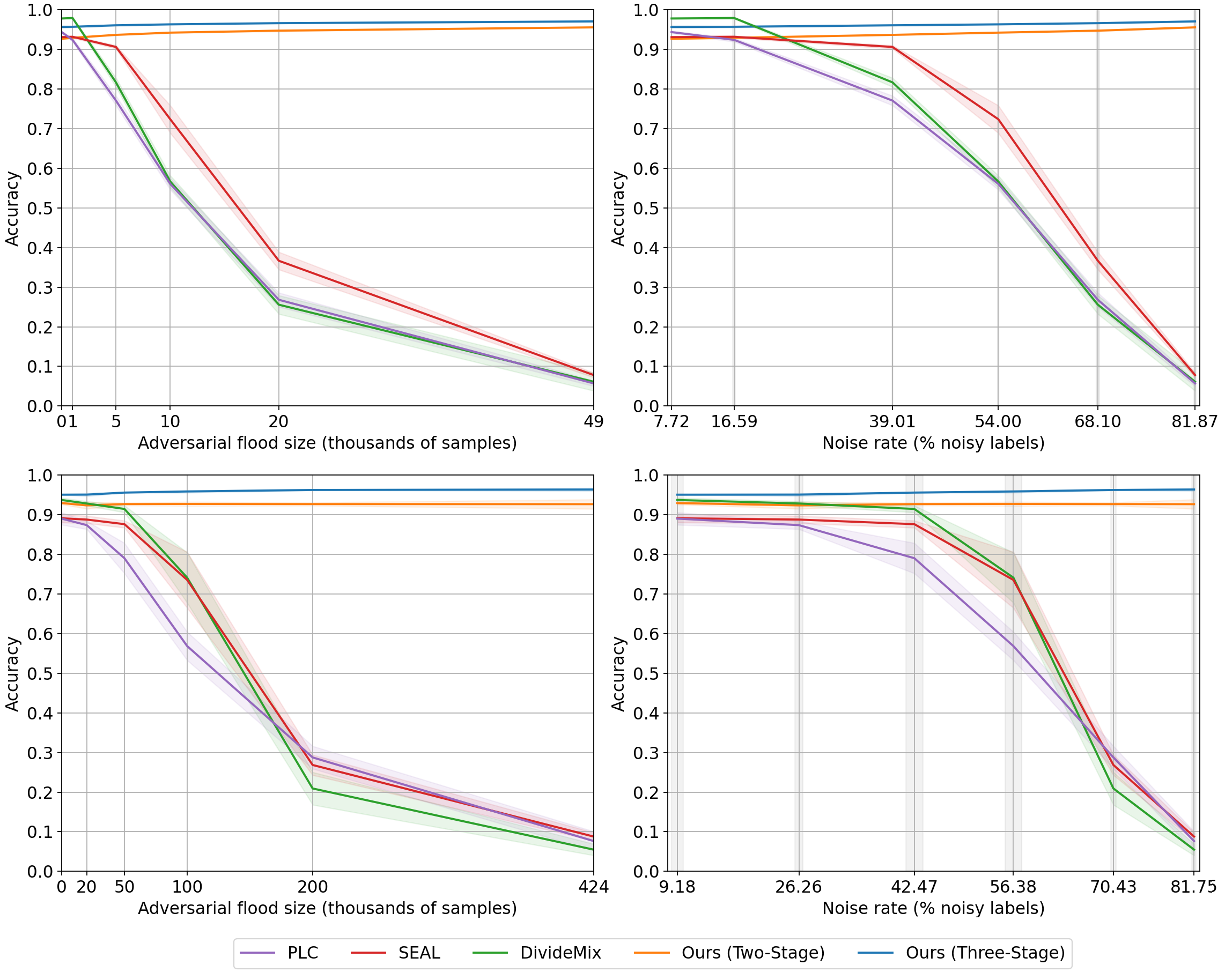}
    \caption{
        Classification accuracy under data flooding attacks evaluated on the MNIST (top row) and SVHN (bottom row) datasets.
        Shaded regions show the 95\% confidence intervals based on the two-sided Student's $t$-test.
        \textbf{Left column:} Horizontal axis scaled to the adversarial attack size.
        \textbf{Right column:} Horizontal axis scaled to the approximate label noise rate. 
        Vertical grey bars indicate standard deviation about the mean of experimental noise rates.
    }   
    \label{fig:flooding-curves}
\end{figure}

For the MNIST dataset, nine good-faith labelers each provided $N_j =$  1,000 labels featuring only natural error, and a single adversarial labeler provided noisy labels varying between 0 and 49,000.
For the SVHN dataset, four good-faith labelers each provided $N_j =$  20,000 labels featuring only natural error, and a single adversarial labeler provided noisy labels varying between 0 and 424,000.
For data flooding experiments, the total amount of data visible to the learner varies with the size of the adversary, with the total amount of good-faith labeled data remaining fixed.

Figure \ref{fig:flooding-curves} shows the classification accuracies of each algorithm in response to increasing amounts of label noise from a single adversary.
We observe that all three state-of-the-art algorithms for learning from noisy labels fail under increasing levels of adversarial noise.
In contrast, our labeler-aware approach remains robust even under extreme adversarial label noise.
These results indicate that recognizing the multiple-labeler paradigm of label gathering is critical in designing robust algorithms for learning from noisy labels. 
We also note the ablative comparisons between the na\"ive application of the DivideMix algorithm, our two-stage framework without stacking, and our three-stage framework that stacks DivideMix on top of our bootstrapped filtered labels.
While labeler-agnostic DivideMix is vulnerable to adversarial noise, and our two-stage labeler-aware framework is robust to adversarial labels but suffers from natural noise, our complete three-stage framework successfully achieves robust performance against both types of noise.

\paragraph{Multiple adversaries}
\begin{figure}
    \centering
    \includegraphics[width=\linewidth]{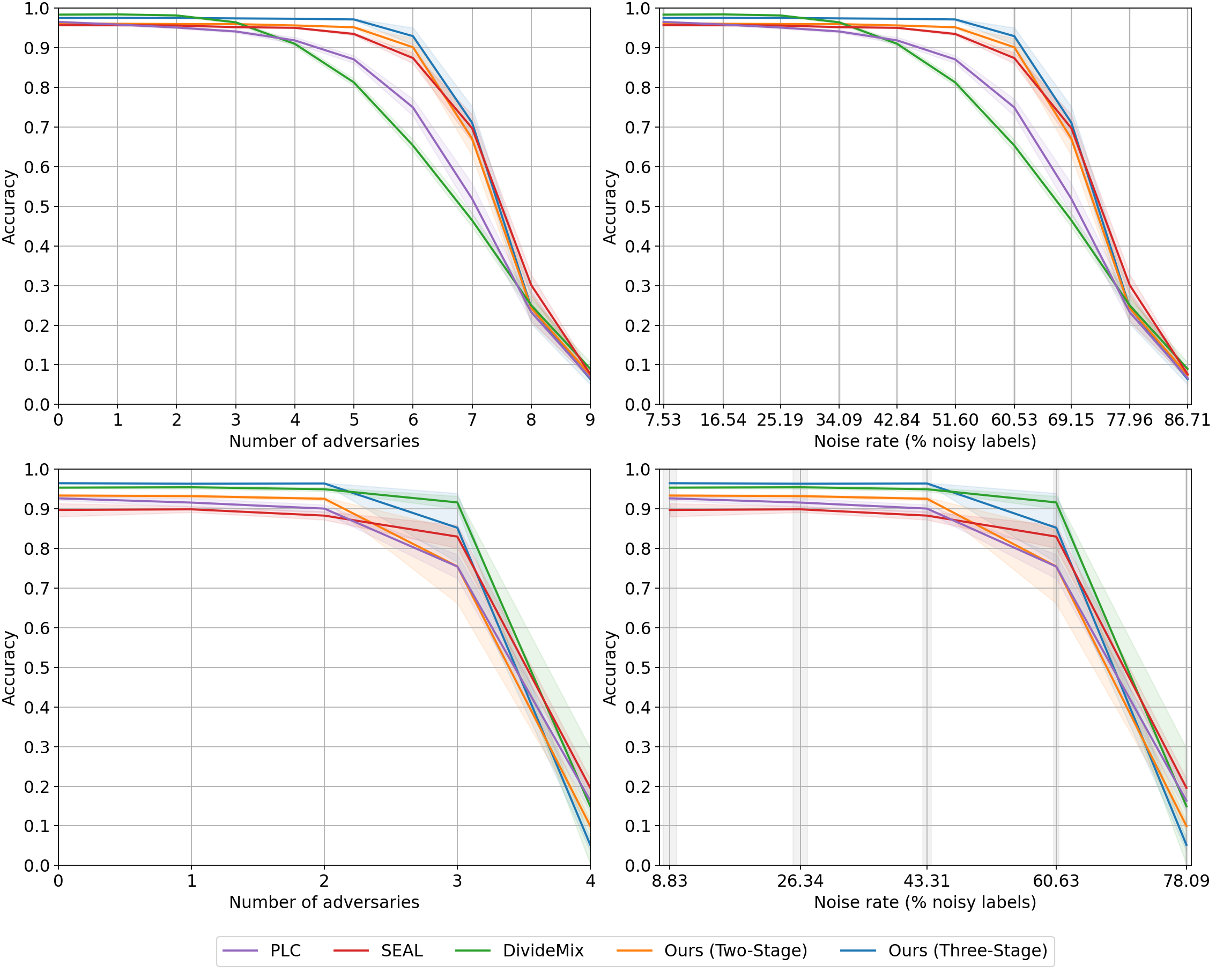}
    \caption{
        Classification accuracy under multiple adversaries attacks evaluated on the MNIST (top row) and SVHN (bottom row) datasets.
        Shaded regions show the 95\% confidence intervals based on the two-sided Student's $t$-test.
        \textbf{Left column:} Horizontal axis scaled to the adversarial attack size.
        \textbf{Right column:} Horizontal axis scaled to the approximate label noise rate. 
        Vertical grey bars indicate standard deviation about the mean of experimental noise rates.
    }   
    \label{fig:multiple-adversaries-curves}
\end{figure}

For the MNIST dataset, each of the ten labelers provided $N_j =$ 5,800 noisy labels, and the number of adversaries $A$ was varied between 0 and 9.
For the SVHN dataset, each of the five labelers provided $N_j = $ 100,000 noisy labels, and $A$ was varied between 0 and 4.
For multiple adversaries experiments, the total amount of data visible to the learner is fixed.

Figure \ref{fig:multiple-adversaries-curves} shows the classification accuracies of each algorithm in response to increasing amounts of adversarial noise caused by multiple adversaries. 
We observe that our labeler-aware framework remains robust against larger fractions of adversarial labelers compared to other methods. 
Naturally, all algorithms, including ours, fail under extreme numbers of adversarial labelers, a well-known phenomenon from ensemble learning that can be traced back to the Condorcet jury theorem \cite{condorcet} and its modern extensions \cite{democratic, condorcet-unreliable, condorcet-extended, condorcet-multiclass}.

\section{Limitations}
\label{sec:limitations}
As discussed above, an important limitation of our labeler-aware framework is its inability to handle extreme fractions of adversarial labelers. 
This difficulty is well-known in the context of learning from crowds \cite{zhong2017quality, whitehill2009whose, kajino2012convex}, and remains an area of open research.

We also note that while our multimodal noisy labels model is more principled than transition matrix-based models, the label noise generation procedure used in our experiments restricts the datasets on which comparisons can be made.
For example, while the CIFAR-10 dataset is a very popular dataset, it is too small in relation to its difficulty to apply our experimental procedure.
This is because there is no amount of labeler training data $\mathcal{D}_j^{tr}$ (drawn without replacement from the full training dataset) that would allow us to train the labeler models $\theta_j$ to a reasonable level of accuracy while still retaining enough data to provide observable labels to the algorithms under test.
Any dataset on which experiments are performed under our labeler-dependent noise model must be either simple enough to learn to a reasonable degree of proficiency on a small subset of the training data (such as the MNIST dataset), or large enough that a greater absolute number of examples could be sampled without replacement to train the labeler models (such as the SVHN dataset).
We note, however, that the severe impact of adversarial noise is still highly damaging to state-of-the-art approaches even on simple datasets, so these datasets remain valuable as testing environments.

Finally, we acknowledge that our multi-stage framework for learning from multiple labelers requires sufficient computational resources.
Stage 1 models may however be trained in parallel, saving on time expenditure; alternatively, an online label gathering environment allows for pipelined training of semi-supervised labeler models as labelers submit their data.

\section{Conclusion}
In this work, we discussed the assumptions and limitations of conventional models of label noise, including transition matrix-based class-conditional noise, instance-dependent noise, and the presence of unreliable or adversarial labelers in the crowdsourcing setting.
We propose a more realistic, labeler-dependent characterization of label noise that integrates both natural error and adversarial labels, acknowledging the reality of multiple-labeler practices for label gathering.
We define two vectors of adversarial label attacks, and propose to evaluate the robustness of noisy label algorithms as their vulnerability to adversarial label noise.
Finally, we present a labeler-aware, multi-stage, modular framework for learning under our multimodal label noise model, and demonstrate our framework's superior robustness against more challenging adversarial noise when compared against state-of-the-art methods for learning from noisy labels.

\section{Broader Impacts}
\label{sec:broader-impacts}
\paragraph{Future research directions}
The ubiquity of label noise in real-world data cannot be overstated, a fact that is reflected in the increasingly large body of work focused on mitigating its effects. 
While conventional models of class-conditional label noise have made great strides, recent work on instance-dependent noise has rightfully pointed out that real-world noise is not necessarily class-conditional.
Our extension of this observation from feature-awareness to labeler-awareness marries the parallel fields of learning from noisy labels and learning from crowds.
This marriage, along with the natural integration of adversarial threats, should form the basis for future research in this critical area.

\paragraph{Adversarial threat modeling}
As with any work proposing consideration of novel attack vectors, the types of attacks described in this work pose serious and potentially destabilizing threats to systems that ignore the threat of adversarial labelers.
There is real-world precedent for the types of attacks we describe: Microsoft's infamous Twitter chatbot Tay \cite{wolf2017we}, its successor Zo \cite{seering2019beyond}, and even commonplace home assistant tools such as Google Home and Amazon Echo \cite{zhang2019defending} have all been subject to malicious data flooding attacks from multiple adversaries.
By codifying these threats and proposing methods for addressing them, our work serves as a preliminary effort in proactively establishing robust defensive measures against adversarial label attacks.

\paragraph{Labeler-dependent biases}
Labeler-dependent biases may exist as natural label noise based on the characteristics of the data on which a labeler learned their beliefs.
In the case of machine labelers, these characteristics will reflect the biases that may be present in the model's training data.
In the case of human labelers, this ``training data'' may take the form of the social, cultural, economic, or educational backgrounds and experiences that inform their beliefs.
Our proposed methods do not leverage these biases, and in fact attempt to minimize the effect of such biases on the downstream classifier.

\paragraph{Practical applications}
We foresee that our work may be of great benefit for industrial applications, where large volumes of noisily-labeled data are commonplace.
We also envision our framework seamlessly integrating into future work on continual or online learning as autonomous agents require the ability to synthesize information from multiple unreliable data sources.

\paragraph{Labeler-aware datasets}
Finally, we hope to initiate an awareness and a shift in how labeled data are gathered, with datasets retaining the information about which data were labeled by which, and by how many, labelers.
With recent work discussing labeling errors in commonly-used benchmark datasets \cite{barz2020we, recht2019imagenet, northcutt2021pervasive, northcutt2021confident}, we are convinced that labeler-aware approaches such as ours are better equipped to more reliably produce accurate labels, identify mislabeled instances, and promote further study and refinement of label quality in standard benchmarks.

\begin{ack}
This work was supported by the U.S. Department of Education, GAANN Grant No. P200A180055.
\end{ack}

{\small
    \bibliographystyle{abbrvnat}
    \bibliography{1_bib.bib}

\begin{thebibliography}{116}
\providecommand{\natexlab}[1]{#1}
\providecommand{\url}[1]{\texttt{#1}}
\expandafter\ifx\csname urlstyle\endcsname\relax
  \providecommand{\doi}[1]{doi: #1}\else
  \providecommand{\doi}{doi: \begingroup \urlstyle{rm}\Url}\fi

\bibitem[Algan and Ulusoy(2021)]{algan2021image}
G.~Algan and I.~Ulusoy.
\newblock Image classification with deep learning in the presence of noisy
  labels: A survey.
\newblock \emph{Knowledge-Based Systems}, 215:\penalty0 106771, 2021.

\bibitem[{Alush} and {Goldberger}(2012)]{ensemble-linear}
A.~{Alush} and J.~{Goldberger}.
\newblock Ensemble segmentation using efficient integer linear programming.
\newblock \emph{IEEE Transactions on Pattern Analysis and Machine
  Intelligence}, 34\penalty0 (10):\penalty0 1966--1977, 2012.
\newblock \doi{10.1109/TPAMI.2011.280}.

\bibitem[Amid et~al.(2019{\natexlab{a}})Amid, Warmuth, and
  Srinivasan]{pmlr-v89-amid19a}
E.~Amid, M.~K. Warmuth, and S.~Srinivasan.
\newblock Two-temperature logistic regression based on the tsallis divergence.
\newblock In \emph{Proceedings of the Twenty-Second International Conference on
  Artificial Intelligence and Statistics}, volume~89 of \emph{Proceedings of
  Machine Learning Research}, pages 2388--2396. PMLR, 16--18 Apr
  2019{\natexlab{a}}.

\bibitem[Amid et~al.(2019{\natexlab{b}})Amid, Warmuth, Anil, and
  Koren]{NEURIPS2019_8cd7775f}
E.~Amid, M.~K.~K. Warmuth, R.~Anil, and T.~Koren.
\newblock Robust bi-tempered logistic loss based on bregman divergences.
\newblock In \emph{Advances in Neural Information Processing Systems},
  volume~32. Curran Associates, Inc., 2019{\natexlab{b}}.

\bibitem[Angluin and Laird(1988)]{angluin1988learning}
D.~Angluin and P.~Laird.
\newblock Learning from noisy examples.
\newblock \emph{Machine Learning}, 2\penalty0 (4):\penalty0 343--370, 1988.

\bibitem[Arazo et~al.(2019)Arazo, Ortego, Albert, O’Connor, and
  McGuinness]{arazo2019unsupervised}
E.~Arazo, D.~Ortego, P.~Albert, N.~O’Connor, and K.~McGuinness.
\newblock Unsupervised label noise modeling and loss correction.
\newblock In \emph{International Conference on Machine Learning}, pages
  312--321. PMLR, 2019.

\bibitem[Barz and Denzler(2020)]{barz2020we}
B.~Barz and J.~Denzler.
\newblock Do we train on test data? purging cifar of near-duplicates.
\newblock \emph{Journal of Imaging}, 6\penalty0 (6):\penalty0 41, 2020.

\bibitem[Bekker and Goldberger(2016)]{bekker2016training}
A.~J. Bekker and J.~Goldberger.
\newblock Training deep neural-networks based on unreliable labels.
\newblock In \emph{2016 IEEE International Conference on Acoustics, Speech and
  Signal Processing (ICASSP)}, pages 2682--2686. IEEE, 2016.

\bibitem[Biggio et~al.(2011)Biggio, Nelson, and Laskov]{biggio2011support}
B.~Biggio, B.~Nelson, and P.~Laskov.
\newblock Support vector machines under adversarial label noise.
\newblock In \emph{Asian Conference on Machine Learning}, pages 97--112. PMLR,
  2011.

\bibitem[Checco et~al.(2020)Checco, Bates, and
  Demartini]{checco2020adversarial}
A.~Checco, J.~Bates, and G.~Demartini.
\newblock Adversarial attacks on crowdsourcing quality control.
\newblock \emph{Journal of Artificial Intelligence Research}, 67:\penalty0
  375--408, 2020.

\bibitem[Chen et~al.(2019)Chen, Liao, Chen, and Zhang]{chen2019understanding}
P.~Chen, B.~B. Liao, G.~Chen, and S.~Zhang.
\newblock Understanding and utilizing deep neural networks trained with noisy
  labels.
\newblock In \emph{International Conference on Machine Learning}, pages
  1062--1070. PMLR, 2019.

\bibitem[Chen et~al.(2020)Chen, Ye, Chen, Zhao, and Heng]{chen2020beyond}
P.~Chen, J.~Ye, G.~Chen, J.~Zhao, and P.-A. Heng.
\newblock Beyond class-conditional assumption: A primary attempt to combat
  instance-dependent label noise.
\newblock \emph{arXiv preprint arXiv:2012.05458}, 2020.

\bibitem[Chen et~al.(2013)Chen, Lin, and Zhou]{chen2013optimistic}
X.~Chen, Q.~Lin, and D.~Zhou.
\newblock Optimistic knowledge gradient policy for optimal budget allocation in
  crowdsourcing.
\newblock In \emph{International Conference on Machine Learning}, pages 64--72.
  PMLR, 2013.

\bibitem[Cheng et~al.(2020)Cheng, Liu, Ramamohanarao, and
  Tao]{cheng2020learning}
J.~Cheng, T.~Liu, K.~Ramamohanarao, and D.~Tao.
\newblock Learning with bounded instance and label-dependent label noise.
\newblock In \emph{International Conference on Machine Learning}, pages
  1789--1799. PMLR, 2020.

\bibitem[da~Silva and Hare(2020)]{da2020humans}
C.~F. da~Silva and T.~A. Hare.
\newblock Humans primarily use model-based inference in the two-stage task.
\newblock \emph{Nature Human Behaviour}, 4\penalty0 (10):\penalty0 1053--1066,
  2020.

\bibitem[Dalvi et~al.(2013)Dalvi, Dasgupta, Kumar, and
  Rastogi]{dalvi2013aggregating}
N.~Dalvi, A.~Dasgupta, R.~Kumar, and V.~Rastogi.
\newblock Aggregating crowdsourced binary ratings.
\newblock In \emph{Proceedings of the 22nd International Conference on World
  Wide Web}, pages 285--294, 2013.

\bibitem[Dawid and Skene(1979)]{dawid1979maximum}
A.~P. Dawid and A.~M. Skene.
\newblock Maximum likelihood estimation of observer error-rates using the em
  algorithm.
\newblock \emph{Journal of the Royal Statistical Society: Series C (Applied
  Statistics)}, 28\penalty0 (1):\penalty0 20--28, 1979.

\bibitem[Dawson and Polikar(2021)]{dawson2021opinionrank}
G.~Dawson and R.~Polikar.
\newblock Opinionrank: Extracting ground truth labels from unreliable expert
  opinions with graph-based spectral ranking.
\newblock \emph{arXiv preprint arXiv:2102.05884}, 2021.

\bibitem[de~Caritat Marquis~de Condorcet(1785)]{condorcet}
M.~J. A.~N. de~Caritat Marquis~de Condorcet.
\newblock \emph{Essai sur l'application de l'analyse {\`a} la probabilit{\'e}
  des d{\'e}cisions rendues {\`a} la pluralit{\'e} des voix}.
\newblock L'imprimerie royale, 1785.

\bibitem[Donmez et~al.(2009)Donmez, Carbonell, and
  Schneider]{donmez2009efficiently}
P.~Donmez, J.~G. Carbonell, and J.~Schneider.
\newblock Efficiently learning the accuracy of labeling sources for selective
  sampling.
\newblock In \emph{Proceedings of the 15th ACM SIGKDD International Conference
  on Knowledge Discovery and Data Mining}, pages 259--268, 2009.

\bibitem[Donmez et~al.(2010)Donmez, Carbonell, and
  Schneider]{donmez2010probabilistic}
P.~Donmez, J.~Carbonell, and J.~Schneider.
\newblock A probabilistic framework to learn from multiple annotators with
  time-varying accuracy.
\newblock In \emph{Proceedings of the 2010 SIAM International Conference on
  Data Mining}, pages 826--837. SIAM, 2010.

\bibitem[Esteva et~al.(2017)Esteva, Kuprel, Novoa, Ko, Swetter, Blau, and
  Thrun]{esteva2017dermatologist}
A.~Esteva, B.~Kuprel, R.~A. Novoa, J.~Ko, S.~M. Swetter, H.~M. Blau, and
  S.~Thrun.
\newblock Dermatologist-level classification of skin cancer with deep neural
  networks.
\newblock \emph{Nature}, 542\penalty0 (7639):\penalty0 115--118, 2017.

\bibitem[Fr{\'e}nay and Verleysen(2013)]{frenay2013classification}
B.~Fr{\'e}nay and M.~Verleysen.
\newblock Classification in the presence of label noise: a survey.
\newblock \emph{IEEE Transactions on Neural Networks and Learning Systems},
  25\penalty0 (5):\penalty0 845--869, 2013.

\bibitem[Frith(2012)]{frith2012role}
C.~D. Frith.
\newblock The role of metacognition in human social interactions.
\newblock \emph{Philosophical Transactions of the Royal Society B: Biological
  Sciences}, 367\penalty0 (1599):\penalty0 2213--2223, 2012.

\bibitem[Garcia et~al.(2019)Garcia, Lehmann, de~Carvalho, and
  Lorena]{garcia2019new}
L.~P. Garcia, J.~Lehmann, A.~C. de~Carvalho, and A.~C. Lorena.
\newblock New label noise injection methods for the evaluation of noise
  filters.
\newblock \emph{Knowledge-Based Systems}, 163:\penalty0 693--704, 2019.

\bibitem[Ghosh et~al.(2011)Ghosh, Kale, and McAfee]{ghosh2011moderates}
A.~Ghosh, S.~Kale, and P.~McAfee.
\newblock Who moderates the moderators? crowdsourcing abuse detection in
  user-generated content.
\newblock In \emph{Proceedings of the 12th ACM Conference on Electronic
  Commerce}, pages 167--176, 2011.

\bibitem[Grofman(1975)]{condorcet-unreliable}
B.~Grofman.
\newblock A comment on ‘democratic theory: A preliminary mathematical
  model.’.
\newblock \emph{Public Choice}, 21\penalty0 (1):\penalty0 99--103, 1975.

\bibitem[Gulshan et~al.(2016)Gulshan, Peng, Coram, Stumpe, Wu, Narayanaswamy,
  Venugopalan, Widner, Madams, Cuadros, et~al.]{gulshan2016development}
V.~Gulshan, L.~Peng, M.~Coram, M.~C. Stumpe, D.~Wu, A.~Narayanaswamy,
  S.~Venugopalan, K.~Widner, T.~Madams, J.~Cuadros, et~al.
\newblock Development and validation of a deep learning algorithm for detection
  of diabetic retinopathy in retinal fundus photographs.
\newblock \emph{JAMA}, 316\penalty0 (22):\penalty0 2402--2410, 2016.

\bibitem[Han et~al.(2018)Han, Yao, Yu, Niu, Xu, Hu, Tsang, and
  Sugiyama]{NEURIPS2018_a19744e2}
B.~Han, Q.~Yao, X.~Yu, G.~Niu, M.~Xu, W.~Hu, I.~Tsang, and M.~Sugiyama.
\newblock Co-teaching: Robust training of deep neural networks with extremely
  noisy labels.
\newblock In \emph{Advances in Neural Information Processing Systems},
  volume~31. Curran Associates, Inc., 2018.

\bibitem[He et~al.(2016{\natexlab{a}})He, Zhang, Ren, and Sun]{he2016deep}
K.~He, X.~Zhang, S.~Ren, and J.~Sun.
\newblock Deep residual learning for image recognition.
\newblock In \emph{Proceedings of the IEEE Conference on Computer Vision and
  Pattern Recognition}, pages 770--778, 2016{\natexlab{a}}.

\bibitem[He et~al.(2016{\natexlab{b}})He, Zhang, Ren, and Sun]{he2016identity}
K.~He, X.~Zhang, S.~Ren, and J.~Sun.
\newblock Identity mappings in deep residual networks.
\newblock In \emph{European conference on computer vision}, pages 630--645.
  Springer, 2016{\natexlab{b}}.

\bibitem[Hickey(1996)]{hickey1996noise}
R.~J. Hickey.
\newblock Noise modelling and evaluating learning from examples.
\newblock \emph{Artificial Intelligence}, 82\penalty0 (1-2):\penalty0 157--179,
  1996.

\bibitem[Huang et~al.(2020)Huang, Zhang, and Zhang]{NEURIPS2020_e0ab531e}
L.~Huang, C.~Zhang, and H.~Zhang.
\newblock Self-adaptive training: beyond empirical risk minimization.
\newblock In \emph{Advances in Neural Information Processing Systems},
  volume~33, pages 19365--19376. Curran Associates, Inc., 2020.

\bibitem[Jagabathula et~al.(2017)Jagabathula, Subramanian, and
  Venkataraman]{jagabathula2017identifying}
S.~Jagabathula, L.~Subramanian, and A.~Venkataraman.
\newblock Identifying unreliable and adversarial workers in crowdsourced
  labeling tasks.
\newblock \emph{The Journal of Machine Learning Research}, 18\penalty0
  (1):\penalty0 3233--3299, 2017.

\bibitem[Joiner et~al.(2017)Joiner, Piva, Turrin, and Chang]{joiner2017social}
J.~Joiner, M.~Piva, C.~Turrin, and S.~W. Chang.
\newblock Social learning through prediction error in the brain.
\newblock \emph{NPJ science of learning}, 2\penalty0 (1):\penalty0 1--9, 2017.

\bibitem[Kajino et~al.(2012)Kajino, Tsuboi, and Kashima]{kajino2012convex}
H.~Kajino, Y.~Tsuboi, and H.~Kashima.
\newblock A convex formulation for learning from crowds.
\newblock In \emph{Proceedings of the AAAI Conference on Artificial
  Intelligence}, volume~26, 2012.

\bibitem[Karimi et~al.(2020)Karimi, Dou, Warfield, and
  Gholipour]{KARIMI2020101759}
D.~Karimi, H.~Dou, S.~K. Warfield, and A.~Gholipour.
\newblock Deep learning with noisy labels: Exploring techniques and remedies in
  medical image analysis.
\newblock \emph{Medical Image Analysis}, 65:\penalty0 101759, 2020.
\newblock ISSN 1361-8415.

\bibitem[Kazmann(1973)]{democratic}
R.~G. Kazmann.
\newblock Democratic organization: A preliminary mathematical model.
\newblock \emph{Public Choice}, 16\penalty0 (1):\penalty0 17--26, 1973.

\bibitem[Khalvati et~al.(2019)Khalvati, Park, Mirbagheri, Philippe, Sestito,
  Dreher, and Rao]{khalvati2019modeling}
K.~Khalvati, S.~A. Park, S.~Mirbagheri, R.~Philippe, M.~Sestito, J.-C. Dreher,
  and R.~P. Rao.
\newblock Modeling other minds: Bayesian inference explains human choices in
  group decision-making.
\newblock \emph{Science Advances}, 5\penalty0 (11):\penalty0 eaax8783, 2019.

\bibitem[Kingma and Ba(2015)]{kingma2015adam}
D.~P. Kingma and J.~Ba.
\newblock Adam: A method for stochastic optimization.
\newblock In \emph{ICLR}, 2015.

\bibitem[Kleindessner and Awasthi(2018)]{pmlr-v80-kleindessner18a}
M.~Kleindessner and P.~Awasthi.
\newblock Crowdsourcing with arbitrary adversaries.
\newblock In \emph{Proceedings of the 35th International Conference on Machine
  Learning}, volume~80 of \emph{Proceedings of Machine Learning Research},
  pages 2708--2717. PMLR, 10--15 Jul 2018.

\bibitem[Kov{\'a}cs et~al.(2010)Kov{\'a}cs, T{\'e}gl{\'a}s, and
  Endress]{kovacs2010social}
{\'A}.~M. Kov{\'a}cs, E.~T{\'e}gl{\'a}s, and A.~D. Endress.
\newblock The social sense: Susceptibility to others’ beliefs in human
  infants and adults.
\newblock \emph{Science}, 330\penalty0 (6012):\penalty0 1830--1834, 2010.

\bibitem[Krizhevsky et~al.(2009)Krizhevsky, Hinton,
  et~al.]{krizhevsky2009learning}
A.~Krizhevsky, G.~Hinton, et~al.
\newblock Learning multiple layers of features from tiny images.
\newblock 2009.

\bibitem[Lachenbruch(1974)]{lachenbruch1974discriminant}
P.~A. Lachenbruch.
\newblock Discriminant analysis when the initial samples are misclassified ii:
  non-random misclassification models.
\newblock \emph{Technometrics}, 16\penalty0 (3):\penalty0 419--424, 1974.

\bibitem[LeCun and Cortes(2010)]{lecun-mnisthandwrittendigit-2010}
Y.~LeCun and C.~Cortes.
\newblock {MNIST} handwritten digit database.
\newblock 2010.

\bibitem[Li et~al.(2019)Li, Wong, Zhao, and Kankanhalli]{li2019learning}
J.~Li, Y.~Wong, Q.~Zhao, and M.~S. Kankanhalli.
\newblock Learning to learn from noisy labeled data.
\newblock In \emph{Proceedings of the IEEE/CVF Conference on Computer Vision
  and Pattern Recognition}, pages 5051--5059, 2019.

\bibitem[Li et~al.(2020)Li, Socher, and Hoi]{li2019dividemix}
J.~Li, R.~Socher, and S.~C. Hoi.
\newblock Dividemix: Learning with noisy labels as semi-supervised learning.
\newblock In \emph{International Conference on Learning Representations}, 2020.

\bibitem[Li et~al.(2017)Li, Wang, Li, Agustsson, and Van~Gool]{li2017webvision}
W.~Li, L.~Wang, W.~Li, E.~Agustsson, and L.~Van~Gool.
\newblock Webvision database: Visual learning and understanding from web data.
\newblock \emph{arXiv preprint arXiv:1708.02862}, 2017.

\bibitem[List and Goodin(2001)]{condorcet-multiclass}
C.~List and R.~E. Goodin.
\newblock Epistemic democracy: Generalizing the condorcet jury theorem.
\newblock \emph{J. Political Philos.}, 9\penalty0 (3):\penalty0 277--306, 2001.
\newblock \doi{https://doi.org/10.1111/1467-9760.00128}.

\bibitem[Liu et~al.(2012)Liu, Peng, and Ihler]{NIPS2012_cd00692c}
Q.~Liu, J.~Peng, and A.~T. Ihler.
\newblock Variational inference for crowdsourcing.
\newblock In \emph{Advances in Neural Information Processing Systems},
  volume~25. Curran Associates, Inc., 2012.

\bibitem[Liu et~al.(2020)Liu, Niles-Weed, Razavian, and
  Fernandez-Granda]{liu2020early}
S.~Liu, J.~Niles-Weed, N.~Razavian, and C.~Fernandez-Granda.
\newblock Early-learning regularization prevents memorization of noisy labels.
\newblock \emph{Advances in Neural Information Processing Systems}, 33, 2020.

\bibitem[Loshchilov and Hutter(2018)]{loshchilov2018decoupled}
I.~Loshchilov and F.~Hutter.
\newblock Decoupled weight decay regularization.
\newblock In \emph{International Conference on Learning Representations}, 2018.

\bibitem[Ma and Olshevsky(2020)]{ma2020adversarial}
Q.~Ma and A.~Olshevsky.
\newblock Adversarial crowdsourcing through robust rank-one matrix completion.
\newblock \emph{Advances in Neural Information Processing Systems}, 33, 2020.

\bibitem[Ma et~al.(2018)Ma, Wang, Houle, Zhou, Erfani, Xia, Wijewickrema, and
  Bailey]{ma2018dimensionality}
X.~Ma, Y.~Wang, M.~E. Houle, S.~Zhou, S.~Erfani, S.~Xia, S.~Wijewickrema, and
  J.~Bailey.
\newblock Dimensionality-driven learning with noisy labels.
\newblock In \emph{International Conference on Machine Learning}, pages
  3355--3364. PMLR, 2018.

\bibitem[Maal{\o}e et~al.(2016)Maal{\o}e, S{\o}nderby, S{\o}nderby, and
  Winther]{maaloe2016auxiliary}
L.~Maal{\o}e, C.~K. S{\o}nderby, S.~K. S{\o}nderby, and O.~Winther.
\newblock Auxiliary deep generative models.
\newblock In \emph{International Conference on Machine Learning}, pages
  1445--1453. PMLR, 2016.

\bibitem[Manwani and Sastry(2013)]{manwani2013noise}
N.~Manwani and P.~Sastry.
\newblock Noise tolerance under risk minimization.
\newblock \emph{IEEE Transactions on Cybernetics}, 43\penalty0 (3):\penalty0
  1146--1151, 2013.

\bibitem[Menon et~al.(2020)Menon, Rawat, Kumar, and Reddi]{48960}
A.~K. Menon, A.~S. Rawat, S.~Kumar, and S.~Reddi.
\newblock Can gradient clipping mitigate label noise?
\newblock In \emph{International Conference on Learning Representations
  (ICLR)}, 2020.

\bibitem[Mirzasoleiman et~al.(2020)Mirzasoleiman, Cao, and
  Leskovec]{NEURIPS2020_8493eeac}
B.~Mirzasoleiman, K.~Cao, and J.~Leskovec.
\newblock Coresets for robust training of deep neural networks against noisy
  labels.
\newblock In \emph{Advances in Neural Information Processing Systems},
  volume~33. Curran Associates, Inc., 2020.

\bibitem[Natarajan et~al.(2013)Natarajan, Dhillon, Ravikumar, and
  Tewari]{NIPS2013_3871bd64}
N.~Natarajan, I.~S. Dhillon, P.~K. Ravikumar, and A.~Tewari.
\newblock Learning with noisy labels.
\newblock In \emph{Advances in Neural Information Processing Systems},
  volume~26. Curran Associates, Inc., 2013.

\bibitem[Netzer et~al.(2011)Netzer, Wang, Coates, Bissacco, Wu, and Ng]{svhn}
Y.~Netzer, T.~Wang, A.~Coates, A.~Bissacco, B.~Wu, and A.~Y. Ng.
\newblock Reading digits in natural images with unsupervised feature learning.
\newblock \emph{NIPS Workshop on Deep Learning and Unsupervised Feature
  Learning}, 2011.

\bibitem[Northcutt et~al.(2021{\natexlab{a}})Northcutt, Jiang, and
  Chuang]{northcutt2021confident}
C.~Northcutt, L.~Jiang, and I.~Chuang.
\newblock Confident learning: Estimating uncertainty in dataset labels.
\newblock \emph{Journal of Artificial Intelligence Research}, 70:\penalty0
  1373--1411, 2021{\natexlab{a}}.

\bibitem[Northcutt et~al.(2021{\natexlab{b}})Northcutt, Athalye, and
  Mueller]{northcutt2021pervasive}
C.~G. Northcutt, A.~Athalye, and J.~Mueller.
\newblock Pervasive label errors in test sets destabilize machine learning
  benchmarks.
\newblock \emph{arXiv preprint arXiv:2103.14749}, 2021{\natexlab{b}}.

\bibitem[Owen et~al.(1989)Owen, Grofman, and Feld]{condorcet-extended}
G.~Owen, B.~Grofman, and S.~L. Feld.
\newblock Proving a distribution-free generalization of the condorcet jury
  theorem.
\newblock \emph{Math. Soc. Sci.}, 17\penalty0 (1):\penalty0 1--16, 1989.

\bibitem[Park et~al.(2017)Park, Go{\"\i}ame, O'Connor, and
  Dreher]{park2017integration}
S.~A. Park, S.~Go{\"\i}ame, D.~A. O'Connor, and J.-C. Dreher.
\newblock Integration of individual and social information for decision-making
  in groups of different sizes.
\newblock \emph{PLoS biology}, 15\penalty0 (6):\penalty0 e2001958, 2017.

\bibitem[Paszke et~al.(2019)Paszke, Gross, Massa, Lerer, Bradbury, Chanan,
  Killeen, Lin, Gimelshein, Antiga, Desmaison, Kopf, Yang, DeVito, Raison,
  Tejani, Chilamkurthy, Steiner, Fang, Bai, and Chintala]{NEURIPS2019_9015}
A.~Paszke, S.~Gross, F.~Massa, A.~Lerer, J.~Bradbury, G.~Chanan, T.~Killeen,
  Z.~Lin, N.~Gimelshein, L.~Antiga, A.~Desmaison, A.~Kopf, E.~Yang, Z.~DeVito,
  M.~Raison, A.~Tejani, S.~Chilamkurthy, B.~Steiner, L.~Fang, J.~Bai, and
  S.~Chintala.
\newblock Pytorch: An imperative style, high-performance deep learning library.
\newblock In \emph{Advances in Neural Information Processing Systems 32}, pages
  8024--8035. Curran Associates, Inc., 2019.

\bibitem[Patrini et~al.(2017)Patrini, Rozza, Krishna~Menon, Nock, and
  Qu]{patrini2017making}
G.~Patrini, A.~Rozza, A.~Krishna~Menon, R.~Nock, and L.~Qu.
\newblock Making deep neural networks robust to label noise: A loss correction
  approach.
\newblock In \emph{Proceedings of the IEEE Conference on Computer Vision and
  Pattern Recognition}, pages 1944--1952, 2017.

\bibitem[Pleiss et~al.(2020)Pleiss, Zhang, Elenberg, and
  Weinberger]{NEURIPS2020_c6102b37}
G.~Pleiss, T.~Zhang, E.~Elenberg, and K.~Q. Weinberger.
\newblock Identifying mislabeled data using the area under the margin ranking.
\newblock In \emph{Advances in Neural Information Processing Systems},
  volume~33, pages 17044--17056. Curran Associates, Inc., 2020.

\bibitem[Raykar et~al.(2009)Raykar, Yu, Zhao, Jerebko, Florin, Valadez, Bogoni,
  and Moy]{raykar2009supervised}
V.~C. Raykar, S.~Yu, L.~H. Zhao, A.~K. Jerebko, C.~Florin, G.~H. Valadez,
  L.~Bogoni, and L.~Moy.
\newblock Supervised learning from multiple experts: whom to trust when
  everyone lies a bit.
\newblock In \emph{ICML}, 2009.

\bibitem[Raykar et~al.(2010)Raykar, Yu, Zhao, Valadez, Florin, Bogoni, and
  Moy]{raykar2010learning}
V.~C. Raykar, S.~Yu, L.~H. Zhao, G.~H. Valadez, C.~Florin, L.~Bogoni, and
  L.~Moy.
\newblock Learning from crowds.
\newblock \emph{Journal of Machine Learning Research}, 11\penalty0 (4), 2010.

\bibitem[Recht et~al.(2019)Recht, Roelofs, Schmidt, and
  Shankar]{recht2019imagenet}
B.~Recht, R.~Roelofs, L.~Schmidt, and V.~Shankar.
\newblock Do imagenet classifiers generalize to imagenet?
\newblock In \emph{International Conference on Machine Learning}, pages
  5389--5400. PMLR, 2019.

\bibitem[Reed et~al.(2015)Reed, Lee, Anguelov, Szegedy, Erhan, and
  Rabinovich]{reed2014training}
S.~E. Reed, H.~Lee, D.~Anguelov, C.~Szegedy, D.~Erhan, and A.~Rabinovich.
\newblock Training deep neural networks on noisy labels with bootstrapping.
\newblock In \emph{ICLR 2015}, 2015.

\bibitem[Reeve et~al.(2019)]{reeve2019classification}
H.~Reeve et~al.
\newblock Classification with unknown class-conditional label noise on
  non-compact feature spaces.
\newblock In \emph{Conference on Learning Theory}, pages 2624--2651. PMLR,
  2019.

\bibitem[Ren et~al.(2018)Ren, Zeng, Yang, and Urtasun]{ren2018learning}
M.~Ren, W.~Zeng, B.~Yang, and R.~Urtasun.
\newblock Learning to reweight examples for robust deep learning.
\newblock In \emph{International Conference on Machine Learning}, pages
  4334--4343. PMLR, 2018.

\bibitem[Rodrigues et~al.(2013)Rodrigues, Pereira, and
  Ribeiro]{rodrigues2013learning}
F.~Rodrigues, F.~Pereira, and B.~Ribeiro.
\newblock Learning from multiple annotators: distinguishing good from random
  labelers.
\newblock \emph{Pattern Recognition Letters}, 34\penalty0 (12):\penalty0
  1428--1436, 2013.

\bibitem[Roman(2009)]{roman2009crowdsourcing}
D.~Roman.
\newblock Crowdsourcing and the question of expertise.
\newblock \emph{Communications of the ACM}, 52\penalty0 (12):\penalty0 12--12,
  2009.

\bibitem[Schroff et~al.(2010)Schroff, Criminisi, and
  Zisserman]{schroff2010harvesting}
F.~Schroff, A.~Criminisi, and A.~Zisserman.
\newblock Harvesting image databases from the web.
\newblock \emph{IEEE Transactions on Pattern Analysis and Machine
  Intelligence}, 33\penalty0 (4):\penalty0 754--766, 2010.

\bibitem[Scott et~al.(2013)Scott, Blanchard, and
  Handy]{scott2013classification}
C.~Scott, G.~Blanchard, and G.~Handy.
\newblock Classification with asymmetric label noise: Consistency and maximal
  denoising.
\newblock In \emph{Conference on Learning Theory}, pages 489--511. PMLR, 2013.

\bibitem[Seering et~al.(2019)Seering, Luria, Kaufman, and
  Hammer]{seering2019beyond}
J.~Seering, M.~Luria, G.~Kaufman, and J.~Hammer.
\newblock Beyond dyadic interactions: Considering chatbots as community
  members.
\newblock In \emph{Proceedings of the 2019 CHI Conference on Human Factors in
  Computing Systems}, pages 1--13, 2019.

\bibitem[Sheng et~al.(2017)Sheng, Zhang, Gu, and Wu]{sheng2017majority}
V.~S. Sheng, J.~Zhang, B.~Gu, and X.~Wu.
\newblock Majority voting and pairing with multiple noisy labeling.
\newblock \emph{IEEE Transactions on Knowledge and Data Engineering},
  31\penalty0 (7):\penalty0 1355--1368, 2017.

\bibitem[Smith and Topin(2019)]{smith2019super}
L.~N. Smith and N.~Topin.
\newblock Super-convergence: Very fast training of neural networks using large
  learning rates.
\newblock In \emph{Artificial Intelligence and Machine Learning for
  Multi-Domain Operations Applications}, volume 11006, page 1100612.
  International Society for Optics and Photonics, 2019.

\bibitem[Snow et~al.(2008)Snow, O'Connor, Jurafsky, and Ng]{snow2008cheap}
R.~Snow, B.~O'Connor, D.~Jurafsky, and A.~Y. Ng.
\newblock Cheap and fast---but is it good? evaluating non-expert annotations
  for natural language tasks.
\newblock In \emph{Proceedings of the Conference on Empirical Methods in
  Natural Language Processing}, EMNLP '08, page 254–263, 2008.

\bibitem[Sohn et~al.(2020)Sohn, Berthelot, Carlini, Zhang, Zhang, Raffel,
  Cubuk, Kurakin, and Li]{NEURIPS2020_06964dce}
K.~Sohn, D.~Berthelot, N.~Carlini, Z.~Zhang, H.~Zhang, C.~A. Raffel, E.~D.
  Cubuk, A.~Kurakin, and C.-L. Li.
\newblock Fixmatch: Simplifying semi-supervised learning with consistency and
  confidence.
\newblock In \emph{Advances in Neural Information Processing Systems},
  volume~33, pages 596--608. Curran Associates, Inc., 2020.

\bibitem[S{\o}nderby et~al.(2016)S{\o}nderby, Raiko, Maal\o~e, S{\o}nderby, and
  Winther]{NIPS2016_6ae07dcb}
C.~K. S{\o}nderby, T.~Raiko, L.~Maal\o~e, S.~K. S{\o}nderby, and O.~Winther.
\newblock Ladder variational autoencoders.
\newblock In \emph{Advances in Neural Information Processing Systems},
  volume~29. Curran Associates, Inc., 2016.

\bibitem[Song et~al.(2020)Song, Kim, Park, Shin, and Lee]{song2020learning}
H.~Song, M.~Kim, D.~Park, Y.~Shin, and J.-G. Lee.
\newblock Learning from noisy labels with deep neural networks: A survey.
\newblock \emph{arXiv preprint arXiv:2007.08199}, 2020.

\bibitem[Sukhbaatar et~al.(2015)Sukhbaatar, Bruna, Paluri, Bourdev, and
  Fergus]{sukhbaatar2015training}
S.~Sukhbaatar, J.~Bruna, M.~Paluri, L.~Bourdev, and R.~Fergus.
\newblock Training convolutional networks with noisy labels.
\newblock In \emph{3rd International Conference on Learning Representations,
  ICLR 2015}, 2015.

\bibitem[Sutskever et~al.(2013)Sutskever, Martens, Dahl, and
  Hinton]{sutskever2013importance}
I.~Sutskever, J.~Martens, G.~Dahl, and G.~Hinton.
\newblock On the importance of initialization and momentum in deep learning.
\newblock In \emph{International conference on machine learning}, pages
  1139--1147. PMLR, 2013.

\bibitem[Tanaka et~al.(2018)Tanaka, Ikami, Yamasaki, and
  Aizawa]{tanaka2018joint}
D.~Tanaka, D.~Ikami, T.~Yamasaki, and K.~Aizawa.
\newblock Joint optimization framework for learning with noisy labels.
\newblock In \emph{Proceedings of the IEEE Conference on Computer Vision and
  Pattern Recognition}, pages 5552--5560, 2018.

\bibitem[Tanno et~al.(2019)Tanno, Saeedi, Sankaranarayanan, Alexander, and
  Silberman]{tanno2019learning}
R.~Tanno, A.~Saeedi, S.~Sankaranarayanan, D.~C. Alexander, and N.~Silberman.
\newblock Learning from noisy labels by regularized estimation of annotator
  confusion.
\newblock In \emph{Proceedings of the IEEE/CVF Conference on Computer Vision
  and Pattern Recognition}, pages 11244--11253, 2019.

\bibitem[Tao et~al.(2020)Tao, Jiang, and Li]{tao2020label}
F.~Tao, L.~Jiang, and C.~Li.
\newblock Label similarity-based weighted soft majority voting and pairing for
  crowdsourcing.
\newblock \emph{Knowledge and Information Systems}, 62:\penalty0 2521--2538,
  2020.

\bibitem[Thulasidasan et~al.(2019)Thulasidasan, Bhattacharya, Bilmes,
  Chennupati, and Mohd-Yusof]{thulasidasan2019combating}
S.~Thulasidasan, T.~Bhattacharya, J.~Bilmes, G.~Chennupati, and J.~Mohd-Yusof.
\newblock Combating label noise in deep learning using abstention.
\newblock In \emph{International Conference on Machine Learning}, pages
  6234--6243. PMLR, 2019.

\bibitem[Vaughan(2017)]{vaughan2017making}
J.~W. Vaughan.
\newblock Making better use of the crowd: How crowdsourcing can advance machine
  learning research.
\newblock \emph{Journal of Machine Learning Research}, 18\penalty0
  (1):\penalty0 7026--7071, 2017.

\bibitem[Wang et~al.(2021)Wang, Han, Liu, Niu, Yang, and
  Gong]{wang2021tackling}
Q.~Wang, B.~Han, T.~Liu, G.~Niu, J.~Yang, and C.~Gong.
\newblock Tackling instance-dependent label noise via a universal probabilistic
  model.
\newblock \emph{arXiv preprint arXiv:2101.05467}, 2021.

\bibitem[Wang et~al.(2020)Wang, Hu, and Hu]{wang2020training}
Z.~Wang, G.~Hu, and Q.~Hu.
\newblock Training noise-robust deep neural networks via meta-learning.
\newblock In \emph{Proceedings of the IEEE/CVF Conference on Computer Vision
  and Pattern Recognition}, pages 4524--4533, 2020.

\bibitem[Wei and Liu(2021)]{wei2020optimizing}
J.~Wei and Y.~Liu.
\newblock When optimizing $ f $-divergence is robust with label noise.
\newblock \emph{ICLR}, 2021.

\bibitem[Welinder et~al.(2010)Welinder, Branson, Perona, and
  Belongie]{welinder2010multidimensional}
P.~Welinder, S.~Branson, P.~Perona, and S.~Belongie.
\newblock The multidimensional wisdom of crowds.
\newblock \emph{Advances in Neural Information Processing Systems},
  23:\penalty0 2424--2432, 2010.

\bibitem[Whitehill et~al.(2009)Whitehill, Wu, Bergsma, Movellan, and
  Ruvolo]{whitehill2009whose}
J.~Whitehill, T.-f. Wu, J.~Bergsma, J.~Movellan, and P.~Ruvolo.
\newblock Whose vote should count more: Optimal integration of labels from
  labelers of unknown expertise.
\newblock \emph{Advances in Neural Information Processing Systems},
  22:\penalty0 2035--2043, 2009.

\bibitem[Wolf et~al.(2017)Wolf, Miller, and Grodzinsky]{wolf2017we}
M.~J. Wolf, K.~W. Miller, and F.~S. Grodzinsky.
\newblock Why we should have seen that coming: comments on microsoft’s tay
  “experiment,” and wider implications.
\newblock \emph{The ORBIT Journal}, 1\penalty0 (2):\penalty0 1--12, 2017.

\bibitem[Wu et~al.(2020)Wu, Zheng, Goswami, Metaxas, and
  Chen]{NEURIPS2020_f4e3ce3e}
P.~Wu, S.~Zheng, M.~Goswami, D.~Metaxas, and C.~Chen.
\newblock A topological filter for learning with label noise.
\newblock In \emph{Advances in Neural Information Processing Systems},
  volume~33, pages 21382--21393. Curran Associates, Inc., 2020.

\bibitem[Xia et~al.(2020)Xia, Liu, Han, Wang, Gong, Liu, Niu, Tao, and
  Sugiyama]{NEURIPS2020_5607fe88}
X.~Xia, T.~Liu, B.~Han, N.~Wang, M.~Gong, H.~Liu, G.~Niu, D.~Tao, and
  M.~Sugiyama.
\newblock Part-dependent label noise: Towards instance-dependent label noise.
\newblock In \emph{Advances in Neural Information Processing Systems},
  volume~33, pages 7597--7610. Curran Associates, Inc., 2020.

\bibitem[Xiao et~al.(2015)Xiao, Xia, Yang, Huang, and Wang]{xiao2015learning}
T.~Xiao, T.~Xia, Y.~Yang, C.~Huang, and X.~Wang.
\newblock Learning from massive noisy labeled data for image classification.
\newblock In \emph{Proceedings of the IEEE Conference on Computer Vision and
  Pattern Recognition}, pages 2691--2699, 2015.

\bibitem[Yan et~al.(2014)Yan, Rosales, Fung, Subramanian, and
  Dy]{yan2014learning}
Y.~Yan, R.~Rosales, G.~Fung, R.~Subramanian, and J.~Dy.
\newblock Learning from multiple annotators with varying expertise.
\newblock \emph{Machine Learning}, 95\penalty0 (3):\penalty0 291--327, 2014.

\bibitem[Yao et~al.(2020)Yao, Liu, Han, Gong, Deng, Niu, and
  Sugiyama]{NEURIPS2020_512c5cad}
Y.~Yao, T.~Liu, B.~Han, M.~Gong, J.~Deng, G.~Niu, and M.~Sugiyama.
\newblock Dual t: Reducing estimation error for transition matrix in
  label-noise learning.
\newblock In \emph{Advances in Neural Information Processing Systems},
  volume~33, pages 7260--7271. Curran Associates, Inc., 2020.

\bibitem[Yi and Wu(2019)]{yi2019probabilistic}
K.~Yi and J.~Wu.
\newblock Probabilistic end-to-end noise correction for learning with noisy
  labels.
\newblock In \emph{Proceedings of the IEEE/CVF Conference on Computer Vision
  and Pattern Recognition}, pages 7017--7025, 2019.

\bibitem[Yu et~al.(2019)Yu, Han, Yao, Niu, Tsang, and Sugiyama]{yu2019does}
X.~Yu, B.~Han, J.~Yao, G.~Niu, I.~Tsang, and M.~Sugiyama.
\newblock How does disagreement help generalization against label corruption?
\newblock In \emph{International Conference on Machine Learning}, pages
  7164--7173. PMLR, 2019.

\bibitem[Zagoruyko and Komodakis(2016)]{Zagoruyko2016WRN}
S.~Zagoruyko and N.~Komodakis.
\newblock Wide residual networks.
\newblock In \emph{BMVC}, 2016.

\bibitem[Zhang et~al.(2017{\natexlab{a}})Zhang, Bengio, Hardt, Recht, and
  Vinyals]{45820}
C.~Zhang, S.~Bengio, M.~Hardt, B.~Recht, and O.~Vinyals.
\newblock Understanding deep learning requires rethinking generalization.
\newblock In \emph{ICLR}, 2017{\natexlab{a}}.

\bibitem[Zhang and Wu(2018)]{zhang2018multi}
J.~Zhang and X.~Wu.
\newblock Multi-label inference for crowdsourcing.
\newblock In \emph{Proceedings of the 24th ACM SIGKDD International Conference
  on Knowledge Discovery \& Data Mining}, pages 2738--2747, 2018.

\bibitem[Zhang et~al.(2016)Zhang, Wu, and Sheng]{zhang2016learning}
J.~Zhang, X.~Wu, and V.~S. Sheng.
\newblock Learning from crowdsourced labeled data: a survey.
\newblock \emph{Artificial Intelligence Review}, 46\penalty0 (4):\penalty0
  543--576, 2016.

\bibitem[Zhang et~al.(2017{\natexlab{b}})Zhang, Sheng, Li, and
  Wu]{zhang2017improving}
J.~Zhang, V.~S. Sheng, T.~Li, and X.~Wu.
\newblock Improving crowdsourced label quality using noise correction.
\newblock \emph{IEEE Transactions on Neural Networks and Learning Systems},
  29\penalty0 (5):\penalty0 1675--1688, 2017{\natexlab{b}}.

\bibitem[Zhang et~al.(2019)Zhang, Zhang, and Zhang]{zhang2019defending}
J.~Zhang, B.~Zhang, and B.~Zhang.
\newblock Defending adversarial attacks on cloud-aided automatic speech
  recognition systems.
\newblock In \emph{Proceedings of the Seventh International Workshop on
  Security in Cloud Computing}, pages 23--31, 2019.

\bibitem[Zhang et~al.(2014)Zhang, Chen, Zhou, and Jordan]{zhang2014spectral}
Y.~Zhang, X.~Chen, D.~Zhou, and M.~I. Jordan.
\newblock Spectral methods meet em: A provably optimal algorithm for
  crowdsourcing.
\newblock \emph{Advances in Neural Information Processing Systems},
  27:\penalty0 1260--1268, 2014.

\bibitem[Zhang et~al.(2021)Zhang, Zheng, Wu, Goswami, and
  Chen]{zhang2021learning}
Y.~Zhang, S.~Zheng, P.~Wu, M.~Goswami, and C.~Chen.
\newblock Learning with feature dependent label noise: a progressive approach.
\newblock \emph{ICLR}, 2021.

\bibitem[Zhang and Sabuncu(2018)]{NEURIPS2018_f2925f97}
Z.~Zhang and M.~Sabuncu.
\newblock Generalized cross entropy loss for training deep neural networks with
  noisy labels.
\newblock In \emph{Advances in Neural Information Processing Systems},
  volume~31. Curran Associates, Inc., 2018.

\bibitem[Zhong et~al.(2017)Zhong, Yang, and Tang]{zhong2017quality}
J.~Zhong, P.~Yang, and K.~Tang.
\newblock A quality-sensitive method for learning from crowds.
\newblock \emph{IEEE Transactions on Knowledge and Data Engineering},
  29\penalty0 (12):\penalty0 2643--2654, 2017.

\bibitem[Zhou et~al.(2014)Zhou, Liu, Platt, and Meek]{zhou2014aggregating}
D.~Zhou, Q.~Liu, J.~C. Platt, and C.~Meek.
\newblock Aggregating ordinal labels from crowds by minimax conditional
  entropy.
\newblock In \emph{International Conference on Machine Learning}, 2014.

\bibitem[Zhu et~al.(2020)Zhu, Liu, and Liu]{zhu2020second}
Z.~Zhu, T.~Liu, and Y.~Liu.
\newblock A second-order approach to learning with instance-dependent label
  noise.
\newblock \emph{arXiv preprint arXiv:2012.11854}, 2020.

\end{thebibliography}
}

\section*{Checklist}
\begin{enumerate}

\item For all authors...
\begin{enumerate}
  \item Do the main claims made in the abstract and introduction accurately reflect the paper's contributions and scope?
    \answerYes{}
  \item Did you describe the limitations of your work?
    \answerYes{See Section~\ref{sec:limitations}.}
  \item Did you discuss any potential negative societal impacts of your work?
    \answerYes{See Section~\ref{sec:broader-impacts}.}
  \item Have you read the ethics review guidelines and ensured that your paper conforms to them?
    \answerYes{}
\end{enumerate}

\item If you are including theoretical results...
\begin{enumerate}
  \item Did you state the full set of assumptions of all theoretical results?
    \answerNA{}
	\item Did you include complete proofs of all theoretical results?
    \answerNA{}
\end{enumerate}

\item If you ran experiments...
\begin{enumerate}
  \item Did you include the code, data, and instructions needed to reproduce the main experimental results (either in the supplemental material or as a URL)?
    \answerYes{See Supplemental Materials.}
  \item Did you specify all the training details (e.g., data splits, hyperparameters, how they were chosen)?
    \answerYes{See Section~\ref{sec:experiments} and Supplemental Materials.}
	\item Did you report error bars (e.g., with respect to the random seed after running experiments multiple times)?
    \answerYes{See Section~\ref{sec:experiments}, we report 95\% confidence intervals for Student's $t$-test over multiple repeated experiments.}
	\item Did you include the total amount of compute and the type of resources used (e.g., type of GPUs, internal cluster, or cloud provider)?
    \answerYes{See Supplemental Materials.}
\end{enumerate}

\item If you are using existing assets (e.g., code, data, models) or curating/releasing new assets...
\begin{enumerate}
  \item If your work uses existing assets, did you cite the creators?
    \answerYes{See Supplemental Materials.}
  \item Did you mention the license of the assets?
    \answerYes{See Supplemental Materials.}
  \item Did you include any new assets either in the supplemental material or as a URL?
    \answerYes{See Supplemental Materials.}
  \item Did you discuss whether and how consent was obtained from people whose data you're using/curating?
    \answerNA{}
  \item Did you discuss whether the data you are using/curating contains personally identifiable information or offensive content?
    \answerNA{}
\end{enumerate}

\item If you used crowdsourcing or conducted research with human subjects...
\begin{enumerate}
  \item Did you include the full text of instructions given to participants and screenshots, if applicable?
    \answerNA{}
  \item Did you describe any potential participant risks, with links to Institutional Review Board (IRB) approvals, if applicable?
    \answerNA{}
  \item Did you include the estimated hourly wage paid to participants and the total amount spent on participant compensation?
    \answerNA{}
\end{enumerate}

\end{enumerate}


\clearpage
\appendix

\section{Additional remarks on labeler-dependent noise}
\subsection{Labeler-dependent noise is a general form of instance-dependent noise}
\label{sec:idn}
Consider instance-dependent noise models that assume a single label noise generation process for the entire training dataset (such as those described in Section 2 of the main paper).
For these models, a noisy label $\hat{y}_i$ for instance $\bm{x}_i$ can be said to be generated following
\setcounter{equation}{1}
\begin{equation}
    \hat{y}_i \sim \mathrm{Cat}(\mathbf{Y}|\bm{\pi}_i)
\end{equation}

where $\bm{\pi}_i = h(\bm{x}_i)$ is an instance-dependent probability distribution characterized by the generating dynamics $h$ of the instance-dependent noise model. 
By adding a labeler dependency following
\begin{equation}
    \hat{y}_{ij} \sim \mathrm{Cat}(\mathbf{Y}|\bm{\pi}_{ij})
\end{equation}

with $\bm{\pi}_{ij} = h_j(\bm{x}_i)$, we obtain the formulation of labeler-dependent noise defined in Section 3 of the main paper. 
Instance-dependent noise of any form can therefore be considered a special case of labeler-dependent noise where $J = 1$.
We note that while for this work we choose $h_j = f_{\theta_j}$, our labeler-dependent noise model holds without loss of generality even if $h$ is chosen to be a different process.

\subsection{On the importance of labeler awareness}
Assume the existence of a theoretical mapping $\mathcal{F}_{\theta^*}: \mathcal{X}^* \mapsto \mathbf{Y}$, where $\mathcal{X}^*$ is the set of all possible data and $\mathbf{Y}$ is the label space, parameterized by an optimal set of parameters $\theta^*$.
Consider that each labeler $j$ generates labels following the parameters $\theta_j$, where $\theta_j$ was learned on an unobserved training dataset $\mathcal{D}^{tr}_j = \{\mathcal{X}^{tr}_j \subsetneq \mathcal{X}^*, \mathcal{Y}^{tr}_j\}$.
$\theta_j$ represents a mapping $\mathcal{F}_{\theta_j}: \mathcal{X}^* \mapsto \mathbf{Y}$ that approximates, but does not equal, $\mathcal{F}_{\theta^*}$.
We observe that for any pair of labelers $\{j, \ell \mid j \neq \ell\}$, under the assumption that $\theta_j \neq \theta_\ell$, the mappings of each labeler will differ, $\mathcal{F}_{\theta_j} \neq \mathcal{F}_{\theta_\ell}$.
We note that differences between $\theta_j$ and $\theta_\ell$ may arise due to different parameterizations, differing $\mathcal{D}^{tr}_j \neq \mathcal{D}^{tr}_\ell$, or both.

Suppose that the data partitions on which each labeler provides data are sampled randomly from $\mathcal{X}^*$ such that $\mathcal{X}_j \neq \mathcal{X}_\ell$ and $\{\mathcal{X}_j, \mathcal{X}_\ell\} \subsetneq \mathcal{X}^*$.
The labels of the data concatenation $\mathcal{X} = \{\mathcal{X}_j, \mathcal{X}_\ell\}$ are therefore generated as a mixture model $\mathcal{F}$, which includes information from both $\mathcal{F}_{\theta_j}$ and $\mathcal{F}_{\theta_\ell}$. 
Without knowledge of which data were labeled by which labeler, the agent being trained on $\mathcal{X}$ is attempting to learn the mixture model $\mathcal{F}$.
Under the assumption that either or both of $j$ or $\ell$ may be adversarial, this mixture model may deviate strongly from $\mathcal{F}_{\theta^*}$, forcing the agent to learn extreme bias.

By retaining information about which labeler provided labels on which data, the agent has the ability to estimate the individual labeler mappings (as in Stage 1 of our proposed framework).
Because each labeler mapping is learned using only the labels provided by that labeler, the damage caused by an adversarial labeler is restricted to only the learner's estimate of the adversary's mapping, thus minimizing the adversary's potential impact.

\subsection{Examples of labeler-dependent noise}
\begin{table}
    \centering
    \caption{Examples of labeler-dependent noisy labels produced by seven good-faith labelers and three adversarial labelers on the MNIST dataset. The labels in each row were produced by the same labeler. Adversarial labels were generated by selecting the second-highest-confidence label based on the adversary's beliefs about the true labels. Correct labels in boldface; incorrect labels marked in red.}
    \begin{tabular}{ccccccccccc}
        \toprule
         &
        \begin{minipage}{.05\textwidth}
            \includegraphics[width=\linewidth]{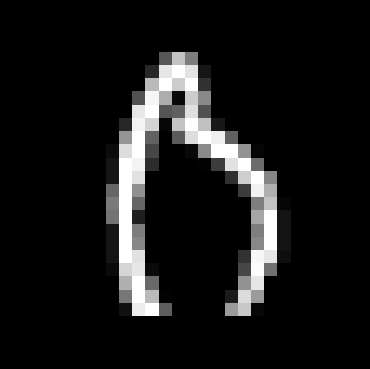}
        \end{minipage} &
        \begin{minipage}{.05\textwidth}
            \includegraphics[width=\linewidth]{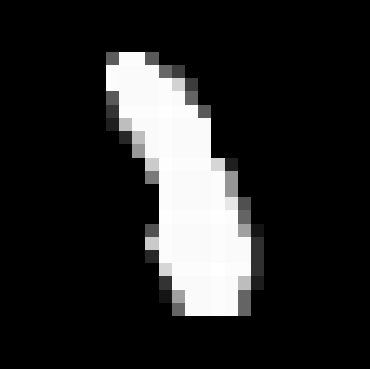}
        \end{minipage} &
        \begin{minipage}{.05\textwidth}
            \includegraphics[width=\linewidth]{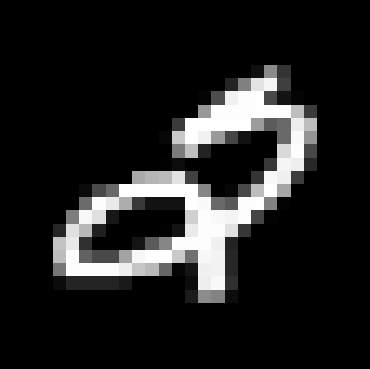}
        \end{minipage} &
        \begin{minipage}{.05\textwidth}
            \includegraphics[width=\linewidth]{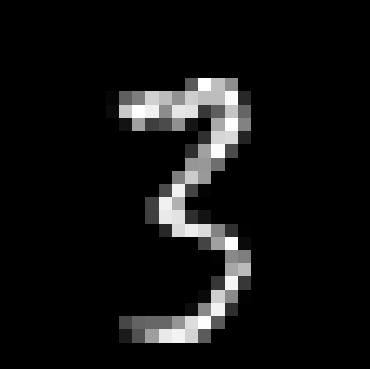}
        \end{minipage} &
        \begin{minipage}{.05\textwidth}
            \includegraphics[width=\linewidth]{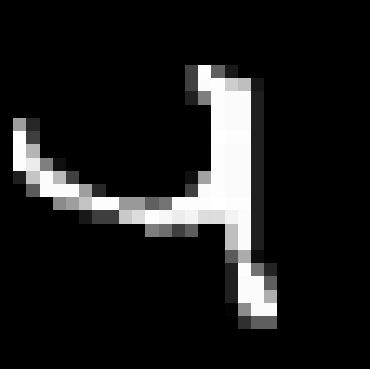}
        \end{minipage} &
        \begin{minipage}{.05\textwidth}
            \includegraphics[width=\linewidth]{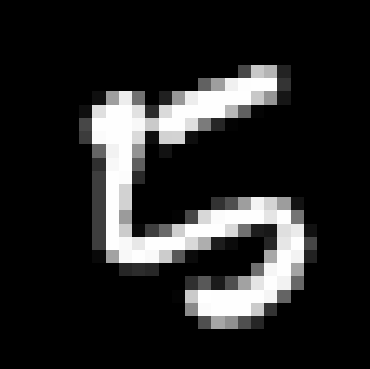}
        \end{minipage} &
        \begin{minipage}{.05\textwidth}
            \includegraphics[width=\linewidth]{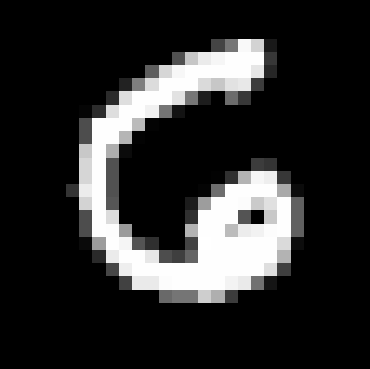}
        \end{minipage} &
        \begin{minipage}{.05\textwidth}
            \includegraphics[width=\linewidth]{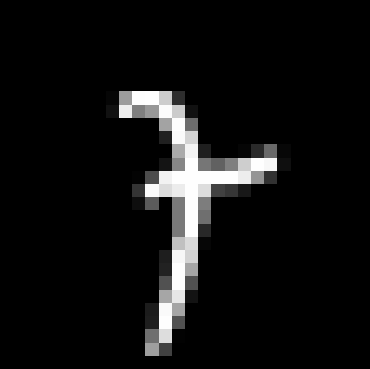}
        \end{minipage} &
        \begin{minipage}{.05\textwidth}
            \includegraphics[width=\linewidth]{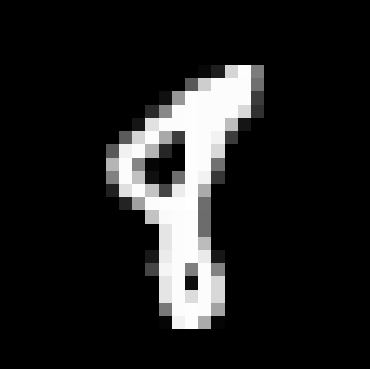}
        \end{minipage} &
        \begin{minipage}{.05\textwidth}
            \includegraphics[width=\linewidth]{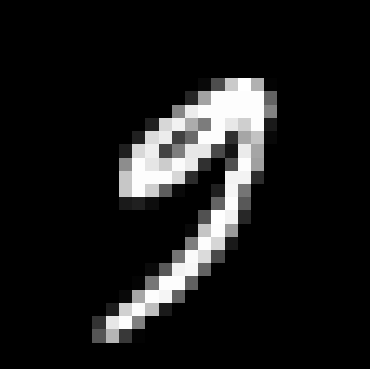}
        \end{minipage} \vspace{1mm} \\
        
        
        
        
        True label & \textbf{0} & \textbf{1} & \textbf{2} & \textbf{3} & \textbf{4} & \textbf{5} & \textbf{6} & \textbf{7} & \textbf{8} & \textbf{9} \\ \midrule
        \multirow{7}{*}{\shortstack{Good-faith\\labelers}} 
        & \textcolor{red}{1} & \textbf{1} & \textcolor{red}{9} & \textbf{3} & \textcolor{red}{9} & \textbf{5} & \textcolor{red}{2} & \textcolor{red}{3} & \textbf{8} & \textbf{9} \\
        & \textbf{0} & \textbf{1} & \textcolor{red}{6} & \textcolor{red}{2} & \textbf{4} & \textcolor{red}{2} & \textcolor{red}{0} & \textcolor{red}{8} & \textbf{8} & \textcolor{red}{4} \\
        & \textcolor{red}{6} & \textcolor{red}{4} & \textcolor{red}{5} & \textcolor{red}{8} & \textbf{4} & \textcolor{red}{6} & \textbf{6} & \textcolor{red}{4} & \textcolor{red}{4} & \textbf{9} \\
        & \textcolor{red}{4} & \textcolor{red}{8} & \textbf{2} & \textbf{3} & \textbf{4} & \textbf{5} & \textcolor{red}{5} & \textbf{7} & \textbf{8} & \textcolor{red}{3} \\
        & \textbf{0} & \textcolor{red}{3} & \textcolor{red}{0} & \textbf{3} & \textcolor{red}{2} & \textcolor{red}{0} & \textbf{6} & \textbf{7} & \textcolor{red}{9} & \textcolor{red}{7} \\
        & \textbf{0} & \textbf{1} & \textbf{2} & \textcolor{red}{7} & \textbf{4} & \textbf{5} & \textcolor{red}{8} & \textcolor{red}{9} & \textcolor{red}{5} & \textbf{9} \\
        & \textcolor{red}{7} & \textcolor{red}{8} & \textbf{2} & \textbf{3} & \textcolor{red}{1} & \textbf{5} & \textbf{6} & \textbf{7} & \textbf{8} & \textcolor{red}{7} \\ \midrule
        \multirow{3}{*}{\shortstack{Adversarial\\labelers}} 
        & \textcolor{red}{5} & \textcolor{red}{8} & \textcolor{red}{9} & \textcolor{red}{2} & \textcolor{red}{2} & \textcolor{red}{7} & \textcolor{red}{0} & \textcolor{red}{5} & \textcolor{red}{9} & \textcolor{red}{5} \\
        & \textcolor{red}{6} & \textcolor{red}{6} & \textcolor{red}{9} & \textcolor{red}{7} & \textcolor{red}{9} & \textcolor{red}{6} & \textcolor{red}{2} & \textcolor{red}{8} & \textcolor{red}{9} & \textcolor{red}{7} \\
        & \textcolor{red}{7} & \textcolor{red}{3} & \textcolor{red}{6} & \textcolor{red}{1} & \textcolor{red}{1} & \textcolor{red}{6} & \textcolor{red}{2} & \textcolor{red}{9} & \textcolor{red}{1} & \textcolor{red}{7} \\
        \bottomrule
    \end{tabular}
    \label{tab:mnist-examples}
\end{table}

\begin{table}
    \centering
    \caption{Examples of labeler-dependent noisy labels produced by seven good-faith labelers and three adversarial labelers on the SVHN dataset. The labels in each row were produced by the same labeler. Adversarial labels were generated by selecting the second-highest-confidence label based on the adversary's beliefs about the true labels. Correct labels in boldface; incorrect labels marked in red.}
    \begin{tabular}{ccccccccccc}
        \toprule
         &
        \begin{minipage}{.05\textwidth}
            \includegraphics[width=\linewidth]{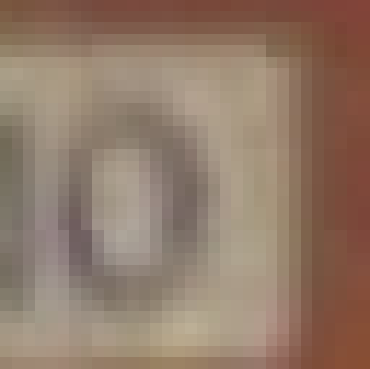}
        \end{minipage} &
        \begin{minipage}{.05\textwidth}
            \includegraphics[width=\linewidth]{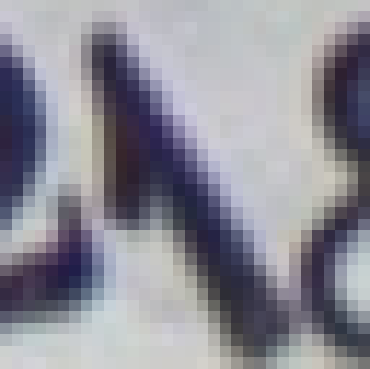}
        \end{minipage} &
        \begin{minipage}{.05\textwidth}
            \includegraphics[width=\linewidth]{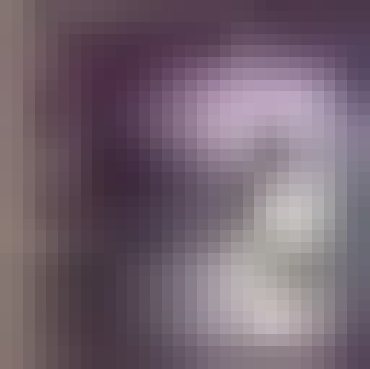}
        \end{minipage} &
        \begin{minipage}{.05\textwidth}
            \includegraphics[width=\linewidth]{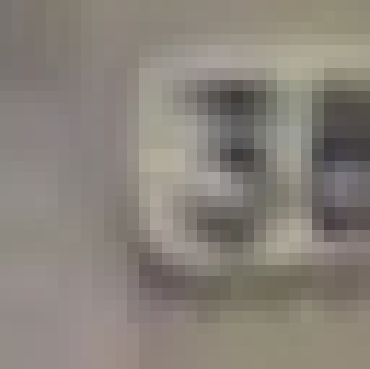}
        \end{minipage} &
        \begin{minipage}{.05\textwidth}
            \includegraphics[width=\linewidth]{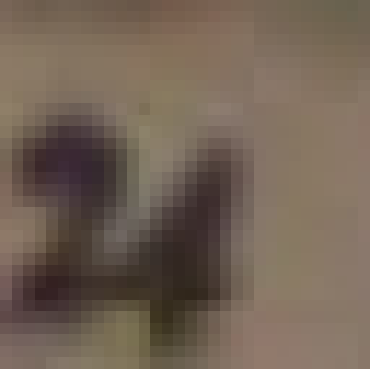}
        \end{minipage} &
        \begin{minipage}{.05\textwidth}
            \includegraphics[width=\linewidth]{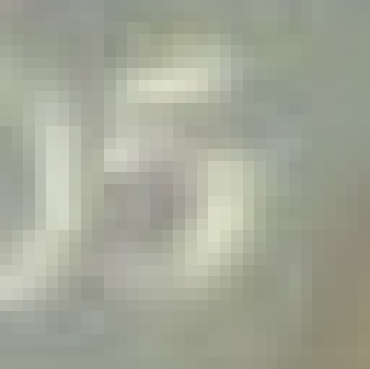}
        \end{minipage} &
        \begin{minipage}{.05\textwidth}
            \includegraphics[width=\linewidth]{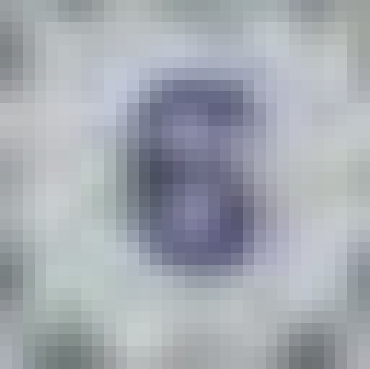}
        \end{minipage} &
        \begin{minipage}{.05\textwidth}
            \includegraphics[width=\linewidth]{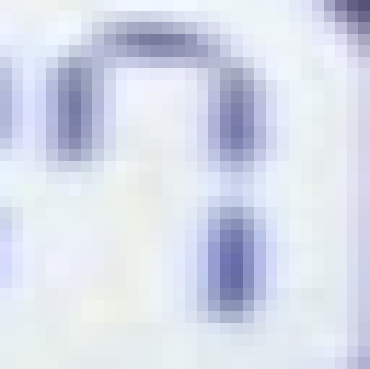}
        \end{minipage} &
        \begin{minipage}{.05\textwidth}
            \includegraphics[width=\linewidth]{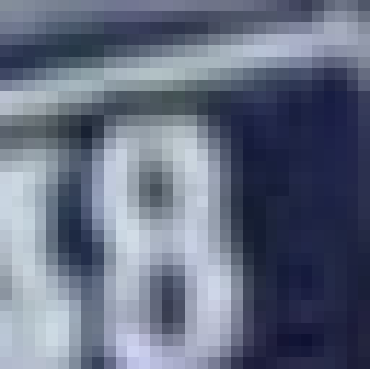}
        \end{minipage} &
        \begin{minipage}{.05\textwidth}
            \includegraphics[width=\linewidth]{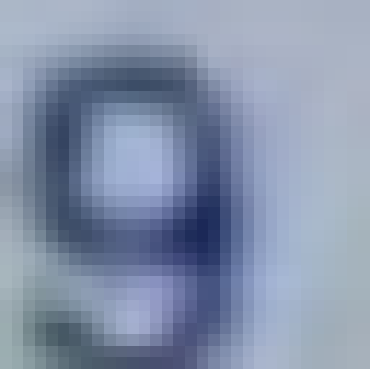}
        \end{minipage} \vspace{1mm} \\
        
        
        
        
        True label & \textbf{0} & \textbf{1} & \textbf{2} & \textbf{3} & \textbf{4} & \textbf{5} & \textbf{6} & \textbf{7} & \textbf{8} & \textbf{9} \\ \midrule
        \multirow{7}{*}{\shortstack{Good-faith\\labelers}} 
        & \textbf{0} & \textbf{1} & \textcolor{red}{9} & \textcolor{red}{9} & \textbf{4} & \textbf{5} & \textbf{6} & \textbf{7} & \textcolor{red}{3} & \textcolor{red}{0} \\
        & \textbf{0} & \textcolor{red}{3} & \textbf{2} & \textbf{3} & \textbf{4} & \textbf{5} & \textbf{6} & \textbf{7} & \textcolor{red}{0} & \textbf{9} \\
        & \textcolor{red}{3} & \textbf{1} & \textcolor{red}{5} & \textcolor{red}{6} & \textcolor{red}{8} & \textcolor{red}{7} & \textcolor{red}{0} & \textbf{7} & \textbf{8} & \textbf{9} \\
        & \textcolor{red}{6} & \textcolor{red}{7} & \textcolor{red}{1} & \textcolor{red}{5} & \textcolor{red}{2} & \textbf{5} & \textbf{6} & \textcolor{red}{3} & \textbf{8} & \textcolor{red}{4} \\
        & \textbf{0} & \textcolor{red}{5} & \textbf{2} & \textbf{3} & \textcolor{red}{7} & \textcolor{red}{9} & \textcolor{red}{8} & \textcolor{red}{8} & \textbf{8} & \textcolor{red}{2} \\
        & \textcolor{red}{5} & \textbf{1} & \textbf{2} & \textbf{3} & \textcolor{red}{0} & \textcolor{red}{0} & \textcolor{red}{9} & \textbf{7} & \textcolor{red}{6} & \textbf{9} \\
        & \textcolor{red}{9} & \textcolor{red}{6} & \textcolor{red}{3} & \textcolor{red}{8} & \textbf{4} & \textcolor{red}{3} & \textcolor{red}{5} & \textcolor{red}{9} & \textcolor{red}{5} & \textcolor{red}{0} \\ \midrule
        \multirow{3}{*}{\shortstack{Adversarial\\labelers}} 
        & \textcolor{red}{6} & \textcolor{red}{8} & \textcolor{red}{7} & \textcolor{red}{2} & \textcolor{red}{2} & \textcolor{red}{3} & \textcolor{red}{4} & \textcolor{red}{2} & \textcolor{red}{1} & \textcolor{red}{4} \\
        & \textcolor{red}{5} & \textcolor{red}{4} & \textcolor{red}{7} & \textcolor{red}{9} & \textcolor{red}{0} & \textcolor{red}{3} & \textcolor{red}{8} & \textcolor{red}{1} & \textcolor{red}{6} & \textcolor{red}{0} \\
        & \textcolor{red}{3} & \textcolor{red}{4} & \textcolor{red}{3} & \textcolor{red}{5} & \textcolor{red}{1} & \textcolor{red}{6} & \textcolor{red}{8} & \textcolor{red}{2} & \textcolor{red}{5} & \textcolor{red}{8} \\
        \bottomrule
    \end{tabular}
    \label{tab:svhn-examples}
\end{table}

Visual examples of instance-dependent noise removing the class-conditional assumption for label noise by generating feature-dependent labels can be found in previous works, such as \cite{chen2020beyond}.
Here, we illustrate labeler-dependent noise as a generalization of instance-dependent noise.
We trained ten labelers identically as described in Appendix \ref{sec:labeler-modeling}, with seven labelers being good-faith and three being adversarial.
In Table \ref{tab:mnist-examples}, we see how the labels generated by each labeler vary widely as a function of both the random initializations as well as the data $\mathcal{X}_j^{tr}$ on which was labeler was trained.
Furthermore, even within just the pool of good-faith labelers, there is a wide variability between labelers with respect to the incorrect label each labeler provides; if label noise were simply instance-dependent, then we would expect to see the same incorrect label provided by all labelers.
We note that the labels provided by adversarial labelers are not random, but rather represent the adversary's best guess about which label will cause maximum confusion to the learning agent; this ``maximum confusion'' approach is illustrated via the coincidence of adversarial labels with erroneous labels provided by good-faith labelers.
Sometimes, however, the adversaries are wrong in their beliefs; in this run, we identified 1168 instances where an adversary would have accidentally provided the correct label.

Table \ref{tab:svhn-examples} shows an example of labels produced on the SVHN dataset following the same procedure as for Table \ref{tab:mnist-examples}.


\section{Visualization and remarks on our three-stage learning framework}
\begin{figure}
    \centering
    \includegraphics[width=\linewidth]{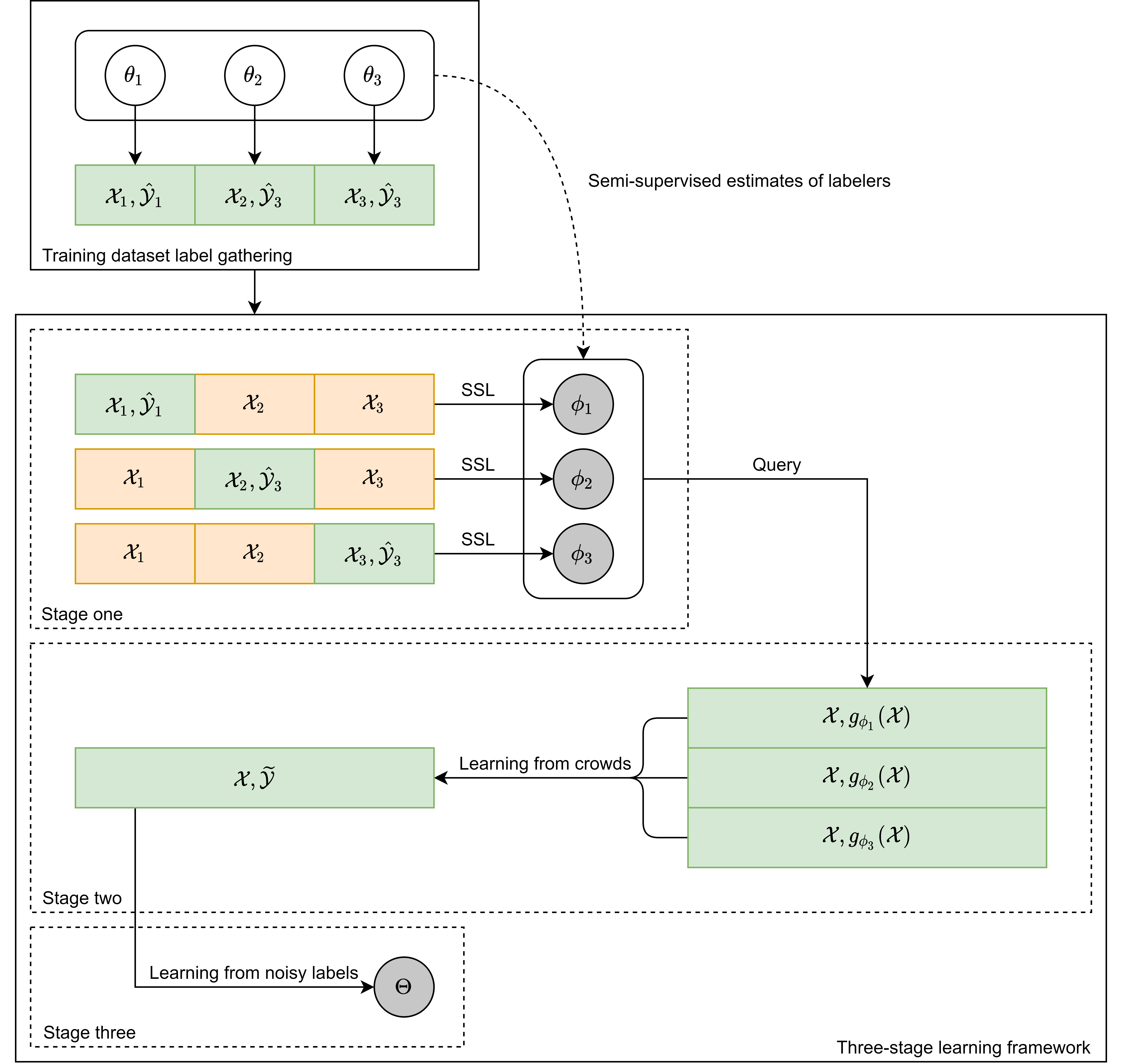}
    \caption{Three-stage learning framework for learning from multiple noisy labelers.}
    \label{fig:framework}
\end{figure}

Our multi-stage learning framework is illustrated in Figure \ref{fig:framework}.
We depict an example with $J = 3$ labelers; the framework extends easily to any number of labelers.
After each labeler provides noisy labels $\hat{\mathcal{Y}}_j$ on their data partitions $\mathcal{X}_j$, in Stage 1 of our framework we construct artificial semi-supervised datasets by treating the labeled data from labelers $\ell \neq j$ as unlabeled.
Each semi-supervised dataset is then used to train models $\phi_j \approx \theta_j$ by exploiting the labels only from labeler $j$.
Each $\phi_j$ is then queried to obtain estimates of what labels $g_{\phi_j}(\mathcal{X})$  labeler $j$ would provide for all available data $\mathcal{X} = \{\mathcal{X}_1, \dots, \mathcal{X}_J\}$.
Note that Stage 1 is completely parallel, and so can be completed efficiently given appropriate resources.
In Stage 2, the full set of $J$ labels for all available data are 
combined into filtered labels $\widetilde{\mathcal{Y}}$ using methods for learning from crowds.
Finally, in Stage 3 we train a classifier $\Theta$ using the filtered (but still noisy) labels from Stage 2 using an algorithm for learning from noisy labels.

Note that for all three stages, there is no prescription on which specific algorithms must be used.
In this sense, our framework is highly modular and adaptive to improvements and advances in each area.
Furthermore, because each stage is self-contained, our framework is able to leverage the theoretical guarantees of each component algorithm achieve the objectives of each stage.

\subsection{Remarks on semi-supervised learning of labeler models}
In Stage 1 of our framework, we estimate the label generation dynamics $\theta_j$ of each labeler $j$ by leveraging semi-supervised learning to train a representative model $\phi_j$.
The labeled data comprises the data-label pairs provided by labeler $j$ directly, and the unlabeled data comprises the data (but not the labels) provided by every other labeler $\ell \neq j$.
A critical consideration is that any label set $\hat{\mathcal{Y}}_j$ may be adversarial, and exhibit extreme label noise; therefore, in this case the $\theta_j$ that is being estimated by $\phi_j$ will be a poor model for learning a classifier.
Our framework explicitly does not attempt to correct this potential label noise in Stage 1, because there is no way to know \textit{a priori} whether a label set $\hat{\mathcal{Y}}_j$ is adversarial.
Therefore, in the presence of adversarial label sources, it is almost surely guaranteed that the corresponding models $\phi_j$ will be poor general classifiers, as the label sets that they produce upon being queried in Stage 2 will exhibit labels of similar quality to those produced by $\theta_j$.

\subsection{Remarks on learning from crowds from queried labels}
Methods for learning from crowds assume that the observed labels are provided directly by each labeler; we are not aware of any literature that attempts to impute missing labels as we do in Stage 1.
We note that our framework subtly violates the parametric assumptions of many methods for learning from crowds by adding an additional layer of joint estimation ($\phi_j$) between the original labeler and the queried labels.
While we have remained intentionally agnostic regarding the parameters $\theta_j$ and $\phi_j$, a considerable drawback of commonly-used approaches such as expectation-maximization \cite{whitehill2009whose, rodrigues2013learning} or Bayesian inference \cite{welinder2010multidimensional} is their dependence on the correct modeling of the precise parameters being estimated. 
For this reason, we suggest using nonparametric approaches, such as weighted majority voting \cite{sheng2017majority, tao2020label} or OpinionRank \cite{dawson2021opinionrank}.

\subsection{Remarks on learning from bootstrapped noisy labels}
In Stage 2, we filter adversarial label noise; in Stage 3, we bootstrap off of the filtered labels $\widetilde{\mathcal{Y}}$ generated in Stage 2 to further filter natural-error label noise. 
This treatment is supported by our empirical results, particularly the ablative comparisons between our framework without the third stage and our complete framework including bootstrapping (Section 5 of the main paper).

\section{Experimental details}
In this section, we report the detailed parameters of each experiment, including the exact parameters of each algorithm.

\subsection{Datasets}
Our experiments investigating algorithmic robustness to data flooding and multiple adversaries attacks are performed on the MNIST \cite{lecun-mnisthandwrittendigit-2010} and SVHN \cite{svhn} datasets.
We would like to acknowledge some other datasets that are commonly used in literature for learning from noisy labels.

The CIFAR-10 dataset \cite{krizhevsky2009learning} comprises 50,000 training images and 10,000 test images in ten classes. 
While we attempted to use the CIFAR-10 dataset for our experiments (following the procedure in Appendix \ref{sec:labeler-modeling}), we found that we could not find an amount of data $\mathcal{D}^{tr}_j$ on which to train each labeler model $\theta_j$ to a reasonable level of accuracy (\textasciitilde 90\%) such that the remaining observable data $\mathcal{D} \setminus \{\bigcup_j \mathcal{D}^{tr}_j\}$ was sufficient to train a classifier.
We consider this inability to be a limitation of our experimental procedure, as the CIFAR-10 dataset is a difficult problem to learn in the low-data regime.
However, we emphasize that this experimental limitation does not extend to the broader impacts of our labeler-aware model, as the severe impact of adversarial label attacks is clearly evident on both the MNIST and SVHN datasets.

The Clothing1M dataset \cite{xiao2015learning} comprises one million images of clothing in 14 classes. 
The images are scraped from websites of fashion retailers, and the labels contain a large, but unknown, amount of label noise.
Because labels for this dataset were gathered using frequency analysis of the text surrounding each image, the dataset does not have any labeler information that can be used to fairly evaluate our proposed learning framework.
Furthermore, because the amount of label noise is unknown, we cannot apply the experimental procedure described in Appendix \ref{sec:labeler-modeling} to perform fair comparisons as the procedure assumes experimental availability of clean ground truth labels..
However, we note that although the attacks described in this work are not directly applicable to this type of label gathering, the methods used to obtain the labels for this dataset are vulnerable to adversarial attacks in the form of false descriptions in the text surrounding each image.
We leave a further investigation of this attack vector for future work.

The WebVision dataset \cite{li2017webvision} comprises approximately 2.4 million images in 1,000 classes.
The images were obtained by crawling from the Flickr website, as well as Google Images search.
The labels for this dataset were assigned as the crawl queries that generated each image. 
Similarly to the Clothing1M dataset, because there is a large, unknown amount of label noise present in the dataset we cannot apply the experimental procedure described in Appendix \ref{sec:labeler-modeling} to perform fair comparisons.
Furthermore, this dataset's label-gathering procedure does not retain information about individual labelers, relying instead on the accuracy of the tagging systems of Flickr and Google Images.
While the authors did use crowdsourcing methods to obtain ``clean'' labels for the validation and test sets, the information about which labelers assigned which labels to which data has been lost.
As with the Clothing1M dataset, we note the vulnerability of this procedure to adversarial false tagging attacks; we leave the study of this method of attack to future work.

\subsection{Labeler modeling}
\label{sec:labeler-modeling}
\setcounter{algorithm}{3}
\begin{algorithm}
    \caption{Multiple-labeler multimodal label noise generation.}
    \begin{algorithmic}[1]
        \label{alg:label-generation}
        \renewcommand{\algorithmicrequire}{\textbf{Inputs:}}
        \renewcommand{\algorithmicensure}{\textbf{Output:}}
        \REQUIRE $J$, the number of label sources; $A$, the number of adversarial labelers; $N_{tr}$, the number of data on which to train each source; $N_j$, the number of data on which labels were provided by each source $j \in J$; $\mathcal{D} = \{\mathcal{X}, \mathcal{Y}\}$, the full training dataset with ground truth labels; $\theta$, the labeler model architecture
        \STATE $\mathcal{D}_{L} \leftarrow$ Initialize the set of noisy label datasets
        \FOR {each labeler $j=1$ to $J$}
            \STATE Sample $\mathcal{D}_j^{tr} \subset \mathcal{D}$ with $|\mathcal{D}_j^{tr}| = N_{tr}$
            \STATE $\mathcal{D} \leftarrow \{\mathcal{D} \setminus \mathcal{D}_j^{tr}\}$ (remove labeler training data from pool)
            \STATE Train $\theta_j$ on $\mathcal{D}_j^{tr} = \{\mathcal{X}_j^{tr}, \mathcal{Y}_j^{tr}\}$
            \STATE Sample $\mathcal{D}_j \subset \mathcal{D}$ with $|\mathcal{D}_j| = N_j$
            \STATE $\mathcal{D} \leftarrow \{\mathcal{D} \setminus \mathcal{D}_j\}$ (remove labeler's provided data from pool)
            \IF {$j \notin A$}
                \STATE $\hat{\mathcal{Y}}_j \leftarrow \mathrm{argmax}_k\ f_{\theta_j}(\mathcal{X}_j)$ (select most confident class as good-faith label)
            \ELSE
                \STATE $\hat{\mathcal{Y}}_j \leftarrow \mathrm{arg}_2\mathrm{max}_k\ f_{\theta_j}(\mathcal{X}_j)$ (select the second most confident class as adversarial label)
            \ENDIF
            \STATE $\mathcal{D}_L \leftarrow \{\mathcal{D}_L \cup \{\mathcal{X}_j, \hat{\mathcal{Y}}_j\}\}$
        \ENDFOR
        \ENSURE $\mathcal{D}_L$
    \end{algorithmic}
\end{algorithm}

For all experiments, labeler-dependent noisy labels were generated following Algorithm \ref{alg:label-generation}.
For the MNIST dataset, each labeler model $\theta_j$ was trained from a randomly-initialized ResNet-18 \cite{he2016deep}.
For the SVHN dataset, each labeler model $\theta_j$ was trained from a randomly-initialized Wide ResNet-50 \cite{Zagoruyko2016WRN}.
Both types of $\theta_j$ architectures were initialized using the implementation provided by PyTorch \cite{NEURIPS2019_9015}.
All $\theta_j$ models were trained for 30 epochs.
All $\theta_j$ models were trained using the AdamW optimizer with default parameters \cite{loshchilov2018decoupled}.
The learning rate for all $\theta_j$ models followed the 1cycle policy \cite{smith2019super} with a maximum learning rate of 0.003.
For the MNIST dataset, the training dataset size for each $\theta_j$ was $N_{tr} = 200$, with a minibatch size of 10.
For the SVHN dataset, the training dataset size for each $\theta_j$ was $N_{tr} = 20,000$, with a minibatch size of 512.
All labeler hyperparameters for both datasets were tuned such that the accuracy of each $\theta_j$ would be approximately 90-95\%. 
We note that because we discard both the $\theta_j$ models as well as the $\mathcal{D}^{tr}_j$ on which they were trained after obtaining their labels $\hat{\mathcal{Y}}_j$, tuning the hyperparameters for the labeler models does not constitute data leakage.

For the MNIST dataset, the total number of labelers was $J=10$, which allowed each labeler to provide a reasonable amount of labels while still retaining a large pool of data for which the adversarial labeler could provide labels in the data flooding experiments.
For the SVHN dataset, the total number of labelers was $J=5$; due to the much larger amount of data required for each $\mathcal{D}^{tr}_j$, a smaller total number of labelers was required such that the adversarial labeler could still flood the dataset in the data flooding experiments.

\subsubsection{Labeler accuracy distributions}
\begin{figure}
    \centering
    \includegraphics[width=\linewidth]{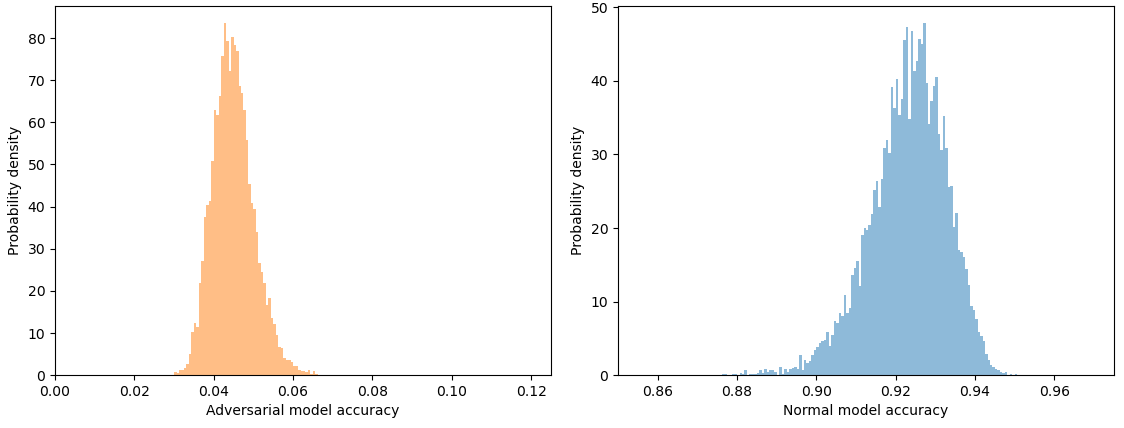}
    \caption{Experimental labeler model accuracy distributions for the MNIST dataset. \textbf{Left:} Accuracy distribution for adversarial labelers. \textbf{Right:} Accuracy distribution for good-faith labelers.}
    \label{fig:mnist-labeler-accs}
\end{figure}

To estimate the consistency of our simulated labelers, we trained 10,000 $\theta_j$ models on random samples from the MNIST training dataset.
Each model was trained on 200 randomly-sampled examples, and we recorded the accuracy of each model as well as its adversarial counterpart on the MNIST test set.
The distributions of the good-faith and adversarial labeler models are shown in Figure \ref{fig:mnist-labeler-accs}.

We observe that even without adversarial labels, the distribution of good-faith labelers trained on very small subsets of the training data do not generalize to 100\% accuracy on the testing data.
Furthermore, there is a relatively wide variance in the accuracy distribution, supporting our hypothesis that label noise from multiple labelers will have more complex behavior than assuming a single labels generation process for the entire training dataset.

\subsection{Implementation details and hyperparameters}
Similarly to \cite{arazo2019unsupervised}, we do not assume the existence of a cleanly-labeled validation set on which to perform hyperparameter tuning or model selection.
Instead, as described below most hyperparameters were kept at the default suggested values found in the publicly-available implementations of each authors' algorithms. 
Generally, we only adjusted hyperparameters to speed up experimental procedures; in each case where this was done, we performed initial training runs to verify that the adjusted hyperparameters were fair.
As in the main paper, we emphasize that due to our lack of hyperparameter fine-tuning, we are likely reporting slightly conservative scores for all algorithms, including our own.
However, we are not interested in squeezing out optimal performance, but rather in obtaining representative evaluations of how each algorithm behaves in response to adversarial label noise.

\subsubsection{Hyperparameters for MNIST experiments}
\begin{table}
    \centering
    \caption{Hyperparameters for MNIST experiments. Parameters marked with a $^*$ follow the recommended settings of the original authors.}
    \begin{tabular}{cll}
        \toprule
        Algorithm & Parameter & Value \\
        \midrule 
        \multirow{9}{*}{PLC \cite{zhang2021learning}} & Base model$^*$ & PreAct ResNet-34 \\
        & Minibatch size$^*$ & 128 \\
        & Optimizer$^*$ & SGD with Nesterov momentum \\
        & Initial LR$^*$ & 0.01 \\
        & LR annealing & 0.1 at epoch 20 \\
        & Momentum$^*$ & 0.9 \\
        & Weight decay$^*$ & 0.0005 \\
        & Warm-up$^*$ & 8 epochs \\
        & Epochs & 30 \\
        \midrule 
        \multirow{7}{*}{SEAL \cite{chen2020beyond}} & Base model$^*$ & Custom CNN$^*$ (Table \ref{tab:seal-cnn}) \\
        & Minibatch size$^*$ & 64 \\
        & Optimizer$^*$ & SGD with momentum \\
        & Initial LR$^*$ & 0.01 \\
        & Momentum$^*$ & 0.5 \\
        & Epochs per iteration$^*$ & 50 \\
        & Iterations$^*$ & 10 \\
        \midrule 
        \multirow{7}{*}{DivideMix \cite{li2019dividemix}} & Base model$^*$ & PreAct ResNet-18 \\
        & Minibatch size & 128 \\
        & Optimizer$^*$ & SGD with momentum \\
        & Initial LR$^*$ & 0.02 \\
        & Momentum$^*$ & 0.9 \\
        & Weight decay$^*$ & 0.0005 \\
        & Epochs & 30 \\
        \midrule 
        \multirow{9}{*}{Auxiliary DGM \cite{maaloe2016auxiliary}} & $\mathrm{dim}(a, z)^*$ & 100 \\
        & $\mathrm{dim}(h)^*$ & 600 \\
        & Minibatch size$^*$ & 200 \\
        & Optimizer$^*$ & Adam \\
        & Initial LR$^*$ & 0.0003 \\
        & $\beta_1, \beta_2$ $^*$ & 0.9, 0.999 \\
        & Classification loss weight$^*$ & $\frac{N_L + N_U}{N_L}$ \\
        & Warm up$^*$ & 200 steps \\
        & Epochs$^*$ & 200 \\
        \bottomrule
    \end{tabular}
    \label{tab:mnist-hyperparams}
\end{table}

All algorithms under test used the same parameters on both the data flooding and multiple adversaries experiments on the MNIST dataset.
All experiments were repeated ten times using uncontrolled random seeds, and our results are reported in the main paper as the mean testing accuracy bounded by the 95\% confidence interval computed using the two-sided Student's $t$-test.
Sources of variability include the randomly-sampled data on which each $\theta_j$ are trained, the random initializations of each network's weights and biases, and any random data augmentation that may be performed as part of any algorithm under test.

A summary of algorithmic hyperparameters is shown in Table \ref{tab:mnist-hyperparams}.

\paragraph{Progressive label correction (PLC)}
For the PLC algorithm \cite{zhang2021learning}, we used the open-source implementation provided by the authors, available at \url{https://github.com/pxiangwu/PLC}.
Following the authors' suggestions, the minibatch size was 128, and the base network was a randomly-initialized PreAct ResNet-34 \cite{he2016identity}.
The optimizer was SGD with Nesterov momentum \cite{sutskever2013importance}, with an initial learning rate of 0.01, momentum of 0.9, and weight decay of 0.0005, and the warm-up period was 8 epochs.
Our implementation differs from that of the original authors by training for 30 epochs, and performing learning rate annealing on the 20th epoch. 
These changes are due to the difference in dataset, and our preliminary experiments showed that continuing to train past 30 epochs did not lead to improved performance.
Therefore, we truncated the training process to conserve computational resources and speed up experiments.

\paragraph{Self-evolution average label (SEAL)}
\begin{table}
    \centering
    \caption{Custom convolutional neural network used by SEAL for the MNIST dataset.}
    \begin{tabular}{ccc}
        \toprule
        Layer & Parameters & Activation\\
        \midrule
        2D convolution & 20 channels, $5 \times 5$ filters & ReLU \\
        2D max pool & $2\times 2$ & -- \\
        2D convolution & 50 channels, $5 \times 5$ filters & ReLU \\
        2D max pool & $2\times 2$ & -- \\
        Flatten & -- & -- \\
        Fully-connected & 500 output features & ReLU \\
        Fully-connected & 10 output features &  Log softmax\\
        \bottomrule
    \end{tabular}
    \label{tab:seal-cnn}
\end{table}
For the SEAL algorithm \cite{chen2020beyond}, we used the open-source implementation provided by the authors, available at \url{https://github.com/chenpf1025/IDN}. 
Because the authors provide their code for training on the MNIST dataset, we used their suggested hyperparameters.
The minibatch size was 64, and the base model was a randomly-initialized custom convolutional neural network (Table \ref{tab:seal-cnn}).
The optimizer was SGD with momentum, with an initial learning rate of 0.01 and momentum of 0.5.
Each iteration was trained for 50 epochs, and the network was trained for 10 iterations.

\paragraph{DivideMix}
\label{sec:mnist_divide}
For the DivideMix algorithm \cite{li2019dividemix}, we used the open-source implementation provided by the authors, available at {\tt https://github.com/LiJunnan1992/DivideMix}. 
Following the authors' suggestions, the base networks were both a randomly-initialized PreAct ResNet-18 \cite{he2016identity}, and the optimizers were SGD with momentum, with an initial learning rate of 0.02, momentum of 0.9, and weight decay of 0.0005.
Our implementation differs from that of the original authors by training for 30 epochs, with a minibatch size of 128.
These changes are due to the difference in dataset, and our preliminary experiments showed that continuing to train past 30 epochs did not lead to improved performance. 
Therefore, we truncated the training process to conserve computational resources and speed up experiments.

\paragraph{Our framework}
For our framework, we require the selection of three component algorithms to use for each of the three stages: \begin{itemize}
    \item We used auxiliary deep generative models \cite{maaloe2016auxiliary} as our semi-supervised learning algorithm for estimating labeler models. 
    We used the open-source implementation available at \url{https://github.com/wohlert/semi-supervised-pytorch}.
    \item We used OpinionRank \cite{dawson2021opinionrank} as our learning from crowds algorithm for integrating redundant labels.
    Our implementation is available in our supplemental code.
    \item We used DivideMix \cite{li2019dividemix} as our algorithm for learning from noisy labels.
    We used the open-source implementation provided by the authors, available at \url{https://github.com/LiJunnan1992/DivideMix}.
\end{itemize}

Auxiliary deep generative models were randomly initialized with $\mathrm{dim}(z) = 100$, $\mathrm{dim}(h) = 600$, and $\mathrm{dim}(a) = 100$, and were trained for 200 epochs with a minibatch size of 200.
The optimizer was Adam \cite{kingma2015adam}, with an initial learning rate of 0.0003 and default $\beta$ parameters of $\beta_1 = 0.9$ and $\beta_2 = 0.999$.
The classification loss weight was $\alpha = \frac{N_L + N_U}{N_L}$, following \cite{maaloe2016auxiliary}.
A deterministic warm-up period of 200 steps was used, following \cite{NIPS2016_6ae07dcb}.

OpinionRank is a nonparametric, deterministic algorithm, and so did not require any hyperparameter selection.

Our usage of DivideMix as the third stage of our framework was implemented identically as described above.

\subsubsection{Hyperparameters for SVHN experiments}
\begin{table}
    \centering
    \caption{Hyperparameters for SVHN experiments. Parameters marked with a $^*$ follow the recommended settings of the original authors.}
    \begin{tabular}{cll}
        \toprule
        Algorithm & Parameter & Value \\
        \midrule 
        \multirow{9}{*}{PLC \cite{zhang2021learning}} & Base model$^*$ & PreAct ResNet-34 \\
        & Minibatch size$^*$ & 128 \\
        & Optimizer$^*$ & SGD with Nesterov momentum \\
        & Initial LR$^*$ & 0.01 \\
        & LR annealing & 0.1 at epoch 20 \\
        & Momentum$^*$ & 0.9 \\
        & Weight decay$^*$ & 0.0005 \\
        & Warm-up$^*$ & 8 epochs \\
        & Epochs & 30 \\
        \midrule 
        \multirow{7}{*}{SEAL \cite{chen2020beyond}} & Base model & ResNet-50 \\
        & Minibatch size & 256 \\
        & Optimizer$^*$ & SGD with momentum \\
        & Initial LR$^*$ & 0.01 \\
        & Momentum$^*$ & 0.5 \\
        & Epochs per iteration$^*$ & 50 \\
        & Iterations$^*$ & 3 \\
        \midrule 
        \multirow{7}{*}{DivideMix \cite{li2019dividemix}} & Base model$^*$ & PreAct ResNet-18 \\
        & Minibatch size & 128 \\
        & Optimizer$^*$ & SGD with momentum \\
        & Initial LR$^*$ & 0.02 \\
        & Momentum$^*$ & 0.9 \\
        & Weight decay$^*$ & 0.0005 \\
        & Epochs & 30 \\
        \midrule 
        \multirow{11}{*}{FixMatch \cite{NEURIPS2020_06964dce}} & Base model$^*$ & Wide ResNet-28 \\
        & Minibatch size$^*$ & 64 \\
        & Optimizer$^*$ & SGD with Nesterov momentum \\
        & Initial LR$^*$ & 0.03 \\
        & Momentum$^*$ & 0.9 \\
        & Weight decay$^*$ & 0.0005 \\
        & EMA decay$^*$ & 0.999 \\
        & Unlabeled minibatch size ratio$^*$ & 7 \\
        & Unlabeled loss coefficient$^*$ & 1.0 \\
        & Pseudo label temperature$^*$ & 0.95 \\
        & Training steps & 9,000 \\
        \bottomrule
    \end{tabular}
    \label{tab:svhn-hyperparams}
\end{table}

All algorithms under test used the same parameters on both the data flooding and multiple adversaries experiments on the SVHN dataset.
All experiments were repeated five times using uncontrolled random seeds, and our results are reported in the main paper as the mean of these results bounded by the 95\% confidence interval computed using the two-sided Student's $t$-test.
Sources of variability include the randomly-sampled data on which each $\theta_j$ are trained, the random initializations of each network's weights and biases, and any random data augmentation that may be performed as part of any algorithm under test.
We combined the standard set of 73,257 training examples with the extra set of 531,131 additional training examples to use as the training dataset $\mathcal{D}$.

A summary of algorithmic hyperparameters is shown in Table \ref{tab:svhn-hyperparams}.

\paragraph{Progressive label correction (PLC)}
For the PLC algorithm \cite{zhang2021learning}, we used the open-source implementation provided by the authors, available at \url{https://github.com/pxiangwu/PLC}.
Following the authors' suggestions, the minibatch size was 128, and the base network was a randomly-initialized PreAct ResNet-34 \cite{he2016identity}.
The optimizer was SGD with Nesterov momentum \cite{sutskever2013importance}, with an initial learning rate of 0.01, momentum of 0.9, and weight decay of 0.0005, and the warm-up period was 8 epochs.
Our implementation differs from that of the original authors by training for  30 epochs, and performing learning rate annealing on the 20th epoch. 
These changes are due to the difference in dataset, and our preliminary experiments showed that continuing to train past 30 epochs did not lead to improved performance.
Therefore, we truncated the training process to conserve computational resources and speed up experiments.

\paragraph{Self-evolution average label (SEAL)}
For the SEAL algorithm \cite{chen2020beyond}, we used the open-source implementation provided by the authors, available at \url{https://github.com/chenpf1025/IDN}. 
Following the authors' suggestions, the optimizer was SGD with momentum, with an initial learning rate of 0.01 and momentum of 0.5.
Our implementation differs from that of the original authors in that the minibatch size was 256, and the base model was a randomly-initialized ResNet-50 \cite{he2016deep}.
Each iteration was trained for 50 epochs, and the network was trained for 3 iterations.
These changes are due to the difference in dataset, and are based on the authors' recommendations for other datasets; our preliminary experiments showed that these hyperparameter settings yielded strong performance in the scenario with no adversarial label noise, so we considered them to be fair for the purposes of our experiments.

\paragraph{DivideMix}
\label{sec:svhn_divide}
For the DivideMix algorithm \cite{li2019dividemix}, we used the open-source implementation provided by the authors, available at {\tt https://github.com/LiJunnan1992/DivideMix}. 
Following the authors' suggestions, the base networks were both a randomly-initialized PreAct ResNet-18 \cite{he2016identity}, and the optimizers were SGD with momentum, with an initial learning rate of 0.02, momentum of 0.9, and weight decay of 0.0005.
Our implementation differs from that of the original authors by training for 30 epochs, with a minibatch size of 128.
These changes are due to the difference in dataset, and our preliminary experiments showed that continuing to train past 30 epochs did not lead to improved performance. 
Therefore, we truncated the training process to conserve computational resources and speed up experiments.

\paragraph{Our framework}
For our framework, we require the selection of three component algorithms to use for each of the three stages: \begin{itemize}
    \item We used FixMatch \cite{NEURIPS2020_06964dce} as our semi-supervised learning algorithm for estimating labeler models. 
    We used the open-source implementation available at \url{https://github.com/kekmodel/FixMatch-pytorch}.
    \item We used OpinionRank \cite{dawson2021opinionrank} as our learning from crowds algorithm for integrating redundant labels.
    Our implementation is available in our supplemental code.
    \item We used DivideMix \cite{li2019dividemix} as our algorithm for learning from noisy labels.
    We used the open-source implementation provided by the authors, available at \url{https://github.com/LiJunnan1992/DivideMix}.
\end{itemize}

The base network for FixMatch was a randomly-initialized Wide Resnet-28 \cite{Zagoruyko2016WRN}.
Following the original authors' suggestions, the minibatch size was 64, the optimizer was SGD with Nesterov momentum \cite{sutskever2013importance} with an initial learning rate of 0.03 and momentum of 0.9, the weight decay was 0.0005, the EMA decay rate was 0.999, the unlabeled minibatch size ratio was 7, the unlabeled loss coefficient was 1.0, and the pseudo label temperature was 0.95.
Our implementation differs from that of the original authors by training for 9,000 steps. 
This change is due to the difference in dataset, and our preliminary experiments showed that continuing to train past 9,000 steps did not lead to improved performance. 
Therefore, we truncated the training process to conserve computational resources and speed up experiments.

OpinionRank is a nonparametric, deterministic algorithm, and so did not require any hyperparameter selection.

Our usage of DivideMix as the third stage of our framework was implemented identically as described above.

\subsection{Hardware}
Each experiment was performed on a single Nvidia Titan RTX on an internal cluster.
Experiments were performed in parallel, as our cluster supports multiple such GPUs.
Our dual-socket cluster is served by two Intel Xeon Gold 5218 2.30 GHz 16-core CPUs.

\subsection{Licenses}
The original authors' DivideMix code is licensed under the terms of the MIT license, available at \url{https://github.com/LiJunnan1992/DivideMix}.
Jesper Wohlert's implementation of auxiliary deep generative models is licensed under the terms of the MIT license, available at \url{https://github.com/wohlert/semi-supervised-pytorch}.
Jungdae Kim's implementation of FixMatch is licensed under the terms of the MIT license, available at \url{https://github.com/kekmodel/FixMatch-pytorch}.

We pledge to release our non-anonymized code (available in anonymized form in these supplemental materials) under the MIT license upon acceptance of this work.

\end{document}